%% file: arxiv_version.tex
\documentclass{article}

\usepackage{arxiv}

\usepackage[utf8]{inputenc} 
\usepackage[T1]{fontenc}    
\usepackage{hyperref}       
\usepackage{url}            
\usepackage{booktabs}       
\usepackage{amsfonts}       
\usepackage{nicefrac}       
\usepackage{microtype}      
\usepackage{lipsum}
\usepackage{amsbsy}

\usepackage{subcaption}
\usepackage{graphicx}

\input{theory_preamule.tex}

\title{A computationally efficient framework for vector representation of persistence diagrams}

\author{
  Kit C~ Chan\\
  Department of Mathematics and Statistics\\
  Bowling Green State University\\
  Bowling Green, OH 43403\\
  \texttt{kchan@bgsu.edu} \\
   \And
  
  Umar ~Islambekov\\
  Department of Mathematics and Statistics\\
  Bowling Green State University\\
  Bowling Green, OH 43403\\
  \texttt{iumar@bgsu.edu} \\
   \And
  Alexey ~Luchinsky \\
  Department of Mathematics and Statistics\\
  Bowling Green State University\\
  Bowling Green, OH 43403\\
  \texttt{aluchi@bgsu.edu} \\
   \And
  Rebecca ~Sanders \\
  Department of Mathematical and Statistical Sciences\\
  Marquette University\\
  Milwaukee, WI 53233\\
  \texttt{rebecca.sanders@marquette.edu} \\
}

\begin{document}
\maketitle

\begin{abstract}
In Topological Data Analysis, a common way of quantifying the shape of data is to use a \emph{persistence diagram} (PD). PDs are multisets of points in $\mathbb{R}^2$ computed using tools of algebraic topology. However, this multi-set structure limits the utility of PDs in applications. Therefore, in recent years efforts have been directed towards extracting informative and efficient summaries from PDs to broaden the scope of their use for machine learning tasks. We propose a computationally efficient framework to convert a PD into a vector in $\mathbb{R}^n$, called a \emph{vectorized persistence block} (VPB). We show that our representation possesses many of the desired properties of vector-based summaries such as stability with respect to input noise, low computational cost and flexibility. Through simulation studies, we demonstrate the effectiveness of VPBs in terms of performance and computational cost within various learning tasks, namely clustering, classification and change point detection.         
\end{abstract}




\clearpage\newpage

\tableofcontents

\section{Introduction}
\input{Introduction}

\section{Related Work}
\label{sec:related_work}
\input{Related_work}



\section{Theory}
\label{sec:theory}
\input{TDA_TheorySectionsFinal}


\section{Simulations}
\label{sec:experiments}
\input{Experiments-2.tex}

\section{Conclusion and future work}
\label{sec:conclusion}
\input{Conclusion}


\end{document}

%% file: theory_preamule.tex
\usepackage{amssymb}
\usepackage{latexsym}
\usepackage{amsthm}
\usepackage{amsfonts}
\usepackage{amscd}
\usepackage{amsxtra}

\newcommand{\R}{\mathbb{R}}

\newcommand{\la}{\lambda}

\providecommand{\gather}{\begin{gather}}
\providecommand{\endgather}{\end{gather}}
\providecommand{\align}{\begin{align}}
\providecommand{\endalign}{\end{align}}
\providecommand{\aligned}{\begin{aligned}}
\providecommand{\endaligned}{\end{aligned}}
\newtheorem{theorem}{Theorem}[section]
\newtheorem{lemma}[theorem]{Lemma}

\newtheorem{corollary}[theorem]{Corollary}

\usepackage{color}
\newcommand{\ALnote}[1]{{\color{red}{AL: #1}}}

\numberwithin{equation}{subsection}

%% file: Introduction.tex
In recent years, topological data analysis (TDA) has become a popular approach to study data that have inherent shape structure \cite{edelsbrunner2010computational,carlsson2009topology}. One of the primary tools of TDA is \emph{persistent homology} (PH) \cite{zomorodian2005computing,edelsbrunner2008persistent}. The idea behind PH is to construct a sequence of nested topological spaces, called \emph{simplicial complexes} (which are indexed by some scale parameter), on top of data points and keep a record of various topological features (such as connected components, one dimensional holes, two dimensional holes, etc.) that emerge and later disappear as the scale parameter increases. A simplicial complex acts as a bridge between discrete data points and the underlying object they are sampled from, and its combinatorial nature makes it suitable for computational purposes. The topological features are detected using the concept of homology from algebraic topology (see e.g. \cite{rotman2013introduction,kaczynski2006computational} for a detailed treatment of the topic). The output of applying PH is a \emph{persistence diagram} (PD) which is a multiset of points in $\mathbb{R}^2$ where each point $(b,d)$ represents a topological feature that is "born" at scale value $b$ and "dies" at scale value $d$. A PD can be viewed as a multiscale topological signature of the data from which it is computed. 

In applications, PDs are often used within statistical and machine learning frameworks for further analysis of the data. A direct way in this regard is to endow the space of PDs with the bottleneck or Wasserstein metrics \cite{mileyko2011probability,cohen2007stability} and use them in distance-based learning settings such as k-medoids or k-nearest neighbors (kNN). Though stability results exist for PDs with respect to these metrics \cite{cohen2007stability,chazal2014persistence,cohen2010lipschitz}, this approach has two limitations: the scope of applicability and the computational cost. Firstly, it is not suitable for machine learning methods requiring input features to be elements of a Hilbert space (the space of PDs can not be turned into a Hilbert space \cite{bubenik2018topological,mileyko2011probability}). Secondly, as the number of points in PDs grows, so does the cost of computing the distance between them. Therefore, in order to facilitate the use of PDs for a broader class of machine learning methods, much research in TDA centers around the problem of extracting useful and computationally efficient summaries from PDs that are identified as vectors of a Hilbert space. There are two common ways to achieve this. The first method to map PDs into a Hilbert space is by defining a kernel function which encodes similarity between PDs and using it in kernel-based machine learning settings \cite{chen2015statistical,kusano2016persistence,li2014persistence,reininghaus2015stable}. The second approach is to extract a vector from a PD which is an element of $\mathbb{R}^n$ \cite{bubenik2015statistical,rieck2017topological,PI,berry2020functional,atienza2020stability,richardson2014efficient,chung2019persistence}. According to \cite{PI}, the desired properties of vectorization-based summaries are stability with respect to input noise, computational efficiency, interpretable connection to PDs and flexibility that allows different treatments of points of PDs based on their relative importance. 

In the present paper, we introduce a computationally efficient framework to extract vectorizaton-based summaries that have all of the above characteristics. First, we switch from the birth-death to birth-persistence coordinates using the map $(b,d)\mapsto (b,p)$, where $p=d-b$. Then, for a given persistence diagram $D$ in the new coordinates, we construct a \emph{persistence block} (PB) which is a function $\tilde{f}(D):\mathbb{R}^2\to \mathbb{R}$ defined as a linear combination of characteristic functions of squares centered at the points of $D$. 
The width of a square centered at point $(b,p)$ depends on the persistence value $p$ and the scale-free parameter $\tau\in[0,1]$. The value of $\tau$ can be set in advance or optimally tuned within the machine learning method in question (see Section \ref{sec:experiments} for more details). Finally, we superimpose a rectangular grid on the domain of the PB $\tilde{f}(D)$ and compute the integral of$\tilde{f}(D)$ with respect to some chosen weight function $w$ over each grid cell. By arranging those integral values into a vector in $\mathbb{R}^n$, where $n$ in the number of grid cells, we obtain a \emph{vectorized persistence block} (VPB). The idea behind a VPB is motivated by the work on \emph{persistence images} (PI) in \cite{PI}. In Section \ref{sec:related_work}, we discuss in more detail the similarities and differences between the two approaches. 

The rest of the paper is organized as follows. 
In Section \ref{sec:theory}, we formally introduce PBs and VPBs and their continuity and stability results with respect to the Wasserstein distance. The results of the simulation studies are presented in Section \ref{sec:experiments}. Finally, in Section \ref{sec:conclusion}, we summarize the contributions of our work and highlight potential directions for future research. Our {\tt R} code to compute VPBs can be found at \hyperlink{https://github.com/uislambekov/TDA-VPB.git}{https://github.com/uislambekov/TDA-VPB.git}.

%% file: Related_work.tex
There are several vectorization-based summaries found in the TDA literature such as persistence landscapes \cite{bubenik2015statistical}, persistence images \cite{PI}, persistence curves \cite{chung2019persistence}, persistence indicator functions \cite{rieck2017topological} and persistent entropy \cite{atienza2020stability}, to name a few. Of these different summaries, the work on persistence images (PI) serves as a motivation for our method. In this section we primarily focus on presenting a detailed discussion about the similarities and differences of the two approaches. 

A PI is obtained by integrating a persistence surface over each cell of a superimposed grid. A persistence surface is a weighted sum of bivariate Gaussian distributions centered at the points of the PD with the covariance matrix $\sigma^2I_{2\times 2}$. The variance $\sigma^2$ of the Gaussians is the same for all points in the PD and is selected by a user. The weights, one for each point, encode a relative importance of the points in the PD. If the birth values are all zero in a PD, as it often happens for the homological dimension $H_0$, then one uses univariate Gaussian distributions instead. In the simulation studies of \cite{PI}, the weights are computed from a piecewise linear function depending only on the persistence value and all the grid cells are of equal size. 

In contrast, for the purposes of our simulation studies, we construct a PB as an unweighted sum of characteristic functions of the squares centered at the points of the PD (see Section \ref{identification map} for more details). The sizes of the squares are not the same but rather change according to their center locations and the parameter $\tau$. The $\tau$ parameter partially corresponds to $\sigma^2$, which controls the spread of the Gaussians in a persistence surface. However, there is an important difference: while $\sigma^2$ is a scale-dependent parameter, $\tau$ is a scale-free parameter with value in $[0,1]$ which makes it easier to select and tune in applications. Our weight function $w:\mathbb{R}^2\rightarrow\mathbb{R}$ is introduced in the vectorization step when we compute the integral of a PB with respect to the weighted area measure $wdA$ over each grid cell. Due to simplicity of construction, one can derive a closed-form formula to compute these integrals for a large class of weight functions which facilitates a fast computation of PVBs in practice. Finally, unlike the grid cells in \cite{PI}, in our simulations they are of different sizes and selected in a data-driven way based on sample percentiles of the birth and persistence values.

%% file: TDA_TheorySectionsFinal.tex


In this section, we provide a mathematical background for our analysis of persistence diagrams. We set forth a foundation and establish continuity and stability results for representing a persistence diagram as a function in an infinite dimensional Hilbert space $L^2$ and as a vector in $\mathbb{R}^n$.  

\subsection{Persistence Block}\label{Persistence Block}

An important aspect of topological data analysis is its ability to track various topological features appearing and disappearing within a filtration built upon data. One approach to code this information is to record the birth $b$ which is the time a topological feature appears, and the death $d$ which is time the exact same feature  disappears. In this way, every topological feature corresponds to exactly one point $(b, d)$ in $\R^2$. Since a topological feature disappears after it appears, we have $0 \le b \le d$, and consequently the point $(b, d)$ lies on or above the line $y=x$. 
An alternative approach is to record the birth $b$ of a topological feature and its persistence $p$, or in other words the life span (i.e. $p=d-b$), of the same feature. Again, each topological feature corresponds to exactly one point $(b, p)$ in $\R^2$. Since the persistence $p$ can be any nonnegative number and if necessary by shifting the birth coordinate values by an appropriate constant, we can assume that the point $(b, p)$ lies in the first quadrant of $\R^2$. For the framework of the present paper, we take the second approach and assume that a persistence diagram (PD) is a collection of points in the first quadrant of $\R^2$, with the interpretation that a point $(b, p)$ in a PD represents the birth and persistence of some topological feature.

Keeping in mind that in practice we only deal with a finite number of PDs and topological features, we can assume all persistence diagrams are finite and there are positive real numbers $\alpha, \beta$ such that all persistence diagrams are subsets of the compact rectangle  
$\Omega = [0, \alpha] \times [0,\beta]$. The set of all finite persistence diagrams is denoted by $P(\Omega)$ and the set of all persistence diagrams with at most $N$ points is denoted by $P_N(\Omega)$.  Technically speaking we can assume  all persistence diagrams are in $P_N(\Omega)$ for some integer $N$.  However, in a careful treatment of our theory dealing with persistence diagrams with different numbers of points, we consider $P_N(\Omega)$ as a subset of $P(\Omega)$ and $P(\Omega) = \bigcup_N P_N(\Omega)$. 

To identify a point in a persistence diagram $D$ in $P(\Omega)$ with a function in some Hilbert space $L^2$, we define a  {\it length function}  $\la : \Omega \to [0,\infty)$, which  is a continuous function such that  $\lambda(x,0) = 0$ whenever $(x,0) \in \Omega$, and 
\begin{eqnarray}\label{kit.0618.1}
0 < \lambda(x,y) < 2y  \ \text{ whenever $(x,y) \in \Omega$ with $ y > 0$.}
\end{eqnarray}
For any point $(x, y) \in \Omega$, let $E(x,y)$ be the square centered at $(x, y)$ with side length  $\la(x, y)$.  That is, 
\begin{eqnarray}\label{0514.1}
 E(x, y) = \left[x-\frac{\la}{2},\,  x+ \frac{\la}{2}\right] \times \left[y-\frac{\la}{2},\,  y+\frac{\la}{2}\right],  \ \text{where $\la = \la(x,y).$}
\end{eqnarray}
Note part of the square $E(x,y)$ may be in the second quadrant, although that part of the square may or may not be included in our integration in Lemma \ref{0617.A} depending on how we apply our theory in the applications.  Regardless, we first study the properties of the whole square $E(x,y)$ in the following lemma, which explains the relationship between the area of overlapping squares and the length function $\lambda$.

\begin{lemma}\label{0511.A}
For any two points $(x,y), (a,b) \in \Omega$, let  $E=E(x,y)$ and $E_0=E(a,b)$ be two squares with side lengths $\la=\lambda(x,y)$ and $\la_0= \lambda(a,b)$, centered at $(x, y)$ and $(a,b)$ respectively. We have the following statements:
\begin{enumerate}
\item[(1)] {\rm Area}$(E\cap E_0) > 0 $ if and only if $ \|(x, y) - (a,b) \|_\infty \  <  \ \frac{\la+\la_0}{2};$   
\item[(2)] {\rm Area}$(E\cup E_0) = \la^2+ {\la}_0^2$ whenever  $ \|(x, y) - (a,b) \|_\infty \ \geq \  \frac{\la+\la_0}{2};$
\item[(3)] If {\rm Area}$(E\cap E_0) > 0 $, then 
\begin{align}
{\rm Area}(E\cap E_0)  & = \bigg( \frac{\la+\la_0}{2}-|x-a|\bigg)\bigg( \frac{\la+\la_0}{2}-|y-b|\bigg)
 \le \left( \frac{\la+\la_0}{2} \right)^2;
\nonumber
\end{align}
\item[(4)]  If {\rm Area}$(E\cap E_0) > 0$, then
\begin{align}
{\rm Area}(E \setminus E_0 ) + {\rm Area}(E_0 \setminus E)
& \le \frac{|\lambda-\lambda_0|^2}{2} + 2(\lambda + \lambda_0) \| (x,y) - (a,b) \|_{\infty}.
\nonumber
\end{align}
\end{enumerate} 
\end{lemma} 

\begin{proof}
The proofs of Statements (1), (2) and (3) follow directly from visualizing two overlapping and two non-overlapping squares. To establish Statement (4), observe that whenever $\mbox{Area}(E \cap E_0) > 0$, we have
\begin{align}
\mbox{Area}(E \setminus E_0)
& = \mbox{Area}(E) - \mbox{Area}(E \cap E_0)
\nonumber \\
& = \lambda^2 -\bigg( \frac{\la+\la_0}{2}-|x-a|\bigg)\bigg( \frac{\la+\la_0}{2}-|y-b|\bigg)
\nonumber \\
& \le \lambda^2 - \left( \frac{\lambda +  \lambda_0}{2} \right)^2 + 
\frac{\lambda + \lambda_0}{2}(|x - a| + |y - b| )
\nonumber \\
& \le \lambda^2 - \left( \frac{\lambda + \lambda_0}{2} \right)^2 + 
(\lambda + \lambda_0) \| (x, y) - (a, b) \|_{\infty}.
\nonumber
\end{align}
By interchanging the roles of $E$ and $E_0$ in the above inequality, we have 
\begin{align}
\mbox{Area}(E_0 \setminus E)
& \le  \lambda_0^2 - \left( \frac{\lambda +  \lambda_0}{2} \right)^2 + 
(\lambda +  \lambda_0) \| (x, y) - (a, b) \|_{\infty}.
\nonumber 
\end{align}
Therefore,
\begin{align}
\mbox{Area}(E \setminus E_0) + \mbox{Area}(E_0 \setminus E)
&  \le  \lambda^2  + \lambda_0^2-  2 \left( \frac{\lambda +  \lambda_0}{2} \right)^2 + 
2 (\lambda + \lambda_0) \| (x, y) - (a, b) \|_{\infty}
\nonumber \\
& = \frac{(\lambda - \lambda_0)^2}{2} +  2 (\lambda +  \lambda_0) \| (x, y) - (a, b) \|_{\infty},
\nonumber
\end{align}
which concludes the proof of Lemma \ref{0511.A}.
\end{proof}

Based on the definition of the length function $\lambda(x,y)$, each square $E(x,y)$ is contained in $[-\beta, \alpha + \beta] \times [0, 2\beta]$, and hence the union $\bigcup_{(x,y) \in \Omega} E(x,y) \subseteq [-\beta, \alpha + \beta] \times [0, 2\beta]$.  By letting $\Omega' = [-\beta, \alpha + \beta] \times [0, 2\beta]$ and letting $w: \Omega' \longrightarrow  [0,\infty)$ be a continuous nonnegative weight function such that 
\begin{align}
w(x,y) > 0, \mbox{ whenever $y > 0$,}
\label{0802.1}
\end{align}
we consider the weighted area measure $wdA$ on $\Omega'$ and its corresponding Hilbert space $L^2(wdA)$.
Any function $f : \Omega \longrightarrow \mathbb{R}$ induces a map $\widetilde{f}: P(\Omega) \longrightarrow L^2(wdA)$ by
\begin{align}\label{0517.1}
\widetilde{f}(D) = \sum_{(a,b) \in D} f(a,b) \chi_{E(a,b)}
\end{align}
where $\chi_{E(a,b)}$ is the characteristic function of the square $E(a,b)$ centered at $(a,b)$ as defined in (\ref{0514.1}).  Since 
$P(\Omega) = \bigcup_{N=1}^{\infty} P_N(\Omega)$, the induced map $\widetilde{f}$ 
can be restricted to $P_N(\Omega)$ for any integer $N \ge 1$. The vector $\widetilde{f}(D)$ in the Hilbert space $L^2(wdA)$ is called a \textit{persistence block}, as we 3-dimensionally visualize a linear combination of  $ \chi_{E(a,b)}$  in  (\ref{0517.1}) as a block. 
The map $\widetilde{f}$ is called the \textit{persistence block transformation induced by $f$}, or simply the \textit{induced persistence block map} if the function $f$ is understood. Taking $\widetilde{f}(D)$ as a function in $L^2(wdA)$, we generally integrate the persistence block $\widetilde{f}(D)$ over the rectangle $\Omega'$. However, there are instances within the theory or within the applications where one may choose to integrate the persistence block $\widetilde{f}(D)$ over the smaller rectangle $\Omega$. In these instances, we only require the nonnegative weight function $w$ to be defined on the rectangle $\Omega$ with $w$ continuously extended to $\Omega'$ as necessary.

In Theorem \ref{0517.A} we  establish a necessary and sufficient condition  for an induced persistence block map $\widetilde{f}$ to be continuous, in the case when $f$ is bounded. For that we need the following lemma, noting that $\Omega$ and $\Omega'$ are compact and $\lambda: \Omega \longrightarrow [0,\infty)$ and $w: \Omega' \longrightarrow [0,\infty)$ are continuous, and hence the functions $\lambda$ and $w$ are necessarily bounded. 

\begin{lemma}\label{0617.A}
Let $\Omega = [0, \alpha] \times [0, \beta]$ and $\Omega' = [-\beta, \alpha+\beta] \times [0, 2\beta]$ with $\alpha, \beta > 0$. Suppose $f: \Omega \longrightarrow \R$ is bounded and $M$ is a positive scalar such that
\begin{align}
\|f\|_\infty \le M, \ \|\lambda\|_\infty  \le M,  \mbox{ and } \|w\|_\infty \le M. 
\label{0617.A.1}
\end{align}
Then we  have the following estimations for any two points $(x,y), (a,b)$ in $\Omega$.
\begin{enumerate}
\item[(1)]  If ${\rm Area}(E(x,y ) \cap E(a ,b )) > 0$, then
\begin{align}
& \ \int_{\Omega} |f(x ,y )\chi_{E(x ,y )} - f(a ,b ) \chi_{E(a ,b )}|^2 wdA
\nonumber \\
\le & \ \int_{\Omega'} |f(x ,y )\chi_{E(x ,y )} - f(a ,b ) \chi_{E(a ,b )}|^2 wdA
\nonumber \\
\le & \  4M^4\bigg( |f(x ,y ) - f(a ,b )| + |\lambda(x ,y ) - \lambda(a ,b )| 
+ \| (x ,y ) - (a , b ) \|_{\infty} \bigg).
\nonumber
\end{align}
\item[(2)]   If ${\rm Area}(E(x ,y ) \cap E(a ,b )) = 0$, then
\begin{align}
\int_{\Omega} |f(x ,y )\chi_{E(x ,y )} - f(a ,b ) \chi_{E(a ,b )}|^2 wdA
& \le \int_{\Omega'} |f(x ,y )\chi_{E(x ,y )} - f(a ,b ) \chi_{E(a ,b )}|^2 wdA
\nonumber \\
& \le 2M^4 \| (x ,y ) - (a , b ) \|_{\infty} .
\nonumber
\end{align}
\end{enumerate}
\end{lemma}

\begin{proof}
For convenience, let $E=E(x , y )$, $\lambda = \lambda(x ,y )$ and $E_0 = E(a ,b )$, $\lambda_0  = \lambda(a , b )$.   To establish the first inequalities in Statements (1) and (2), note that $\Omega \subseteq \Omega'$ and the integrand for the two integrals is nonnegative.

Focusing on the second half of the inequalities, observe that
\begin{align}
\left| f(x , y )  -  f(a , b )  \right|^2 
& \le ( |f(x , y )| + |f(a , b )|)  \left| f(x , y )  -  f(a , b )  \right|
\label{0617.A.3} \\
& \le 2M \left| f(x , y )  -  f(a , b )  \right|.
\nonumber
\end{align}
By repeated applications of the inequalities in (\ref{0617.A.1}), we have
\begin{align}
& \int_{\Omega'} \left| f(x , y ) \chi_{E} - f(a , b ) \chi_{E_0} \right|^2 w dA
\label{0617.A.2} \\
= & \int_{E \cap E_0} \left| f(x , y )  -  f(a , b )  \right|^2 w dA
+ \int_{E \setminus E_0} \left| f(x , y )  \right|^2 w dA
+ \int_{E_0 \setminus E} \left| f(a , b )  \right|^2 w dA
\nonumber \\
\le & \ \left| f(x , y )  -  f(a , b )  \right|^2 \int_{E \cap E_0} wdA + M^2 \int_{E \setminus E_0} wdA + 
M^2 \int_{E_0 \setminus E} wdA
\nonumber \\
\le & \ M \left| f(x , y )  -  f(a , b )  \right|^2 \mbox{Area}( E\cap E_0 ) + M^3 \mbox{Area}(E \setminus E_0) +
M^3 \mbox{Area}(E_0 \setminus E)
\nonumber \\
\le & \ 2M^2 \left| f(x , y )  -  f(a , b )  \right|  \mbox{Area}( E\cap E_0 ) + M^3\bigg(\mbox{Area}( E\setminus E_0 ) + \mbox{Area}( E_0\setminus E)\bigg),  \ \text{by (\ref{0617.A.3}).}
\nonumber 
\end{align}

If $\mbox{Area}(E \cap E_0) > 0$, by Statement (3) of Lemma \ref{0511.A}, we have
\begin{align}
\mbox{Area}(E \cap E_0) 
& \le \bigg(\frac{\lambda + \lambda_0}{2}\bigg)^2 \le M^2, \mbox{ by (\ref{0617.A.1})}. 
\label{0617.A.4}
\end{align}
Similarly, by Statement (4) of Lemma \ref{0511.A}, we have 
\begin{align}
\mbox{Area}( E\setminus E_0 ) + \mbox{Area}( E_0\setminus E)
& \le \frac{|\lambda - \lambda_0|^2}{2} + 2(\lambda + \lambda_0) \| (x , y ) - (a , b ) \|_{\infty}
\label{0617.A.5} \\
& \le \frac{(\lambda + \lambda_0) |\lambda - \lambda_0|}{2} + 2(\lambda + \lambda_0) \| (x , y ) - (a , b ) \|_{\infty}
\nonumber \\
& \le M |\lambda - \lambda_0| + 4M \| (x , y ) - (a , b ) \|_{\infty}, \mbox{ by (\ref{0617.A.1})}
\nonumber \\
& = 4M \bigg(|\lambda - \lambda_0| + \| (x , y ) - (a , b ) \|_{\infty}\bigg).
\nonumber
\end{align}
Substituting  (\ref{0617.A.4}) and (\ref{0617.A.5}) into (\ref{0617.A.2}) yields
\begin{align}
& \ \int_{\Omega'} \left| f(x , y ) \chi_{E} - f(a , b ) \chi_{E_0} \right|^2 w dA
\label{0617.A.6} \\
\le & \ 2M^4 \left| f(x , y )  -  f(a , b )  \right| + 4M^4\bigg( |\lambda - \lambda_0| + \| (x , y ) - (a , b ) \|_{\infty}\bigg)
\nonumber \\
= & \ 4M^4\bigg( \left| f(x , y )  -  f(a , b )  \right| + |\lambda - \lambda_0| + \| (x , y ) - (a , b ) \|_{\infty}\bigg).
\nonumber
\end{align}

If $\mbox{Area}(E \cap E_0) = 0$, then
\begin{align}
\mbox{Area}( E\setminus E_0 ) + \mbox{Area}( E_0 \setminus E)
&  =  \mbox{Area}(E) + \mbox{Area}(E_0) 
\label{0617.A.7} \\  
& =  \  \lambda^2 + \lambda_0^2 
\nonumber \\
& \le  M \lambda + M \lambda_0, \mbox{ by (\ref{0617.A.1})}
\nonumber \\
& =  M( \lambda +  \lambda_0) 
\nonumber \\ 
& \le  2M \| (x , y ) - (a , b ) \|_{\infty}, \text{ by Statement (2) of Lemma \ref{0511.A}}.
\nonumber 
\end{align}
Combining the above inequality with (\ref{0617.A.2}), we get
\begin{align}
& \ \int_{\Omega'} \left| f(x , y ) \chi_{E} - f(a , b ) \chi_{E_0} \right|^2 w dA
\nonumber \\
\le & \ M^3(\mbox{Area}( E\setminus E_0 ) + \mbox{Area}( E_0\setminus E))
\nonumber \\
\le & \ 2M^4 \| (x , y ) - (a , b ) \|_{\infty},
\nonumber
\end{align}
which concludes the proof of the lemma.
\end{proof}

In order to establish a necessary and sufficient condition to ensure the continuity of an induced persistence block map, we utilize the $p$-Wasserstein distance for the metric on the space of persistence diagrams.
For $1 \le p < \infty$, the {\it $p$-Wasserstein distance}  $W_p(D, D_0) $ between two persistence diagrams $D, D_0$ in $P_N(\Omega)$ is defined by
\begin{align}
W_p(D, D_0) = 
\inf \left\{ \biggl( \sum_{(a,b) \in D} \| \gamma(a,b) - (a,b) \|_{\infty}^p \biggr)^{1/p} \bigg| \mbox{ $\gamma: D \longrightarrow D_0$ is a bijection} \right\}.
\label{pWasserdef1}
\end{align}
where additional points along the $x$-axis may be added to $D$ or $D_0$ in order to define a bijection $\gamma$.  For $0 < p < 1$, we similarly define the {\it $p$-Wasserstein distance} $W_p(D, D_0)$ by
\begin{align}
W_p(D, D_0) = 
\inf \left\{\sum_{(a,b) \in D} \| \gamma(a,b) - (a,b) \|_{\infty}^p \bigg| \mbox{ $\gamma: D \longrightarrow D_0$ is a bijection} \right\},
\label{pWasserdef2}
\end{align}
To explain the difference between the two formulas in (\ref{pWasserdef1}) and (\ref{pWasserdef2}), let 
$x = (x_1, x_2, \dots, x_n)$ and $y = (y_1, y_2, \dots, y_n)$ be two vectors in $\mathbb{R}^n$.  The formula
$d_p(x,y) = \left( \sum_{j=1}^{n}|x_j - y_j|^p \right)^{1/p}$ defines a metric on $\mathbb{R}^n$ only when $1 \le p < \infty$.  However, in the case when $0 < p < 1$, the above formula fails to define a metric.  On the positive side, by modifying the formula to $d_p(x,y) = \sum_{j=1}^{n}|x_j - y_j|^p$, we generate a metric on $\mathbb{R}^n$; see Rudin \cite[Section 1.47, page 36]{RUDIN}.  Thus in both (\ref{pWasserdef1}) and (\ref{pWasserdef2}), we take the infimum of a metric.

Before we state the main theorem of the current subsection, we remark that the weight function $w$ is defined on $\Omega'$, and hence the norm of the Hilbert space $L^2(wdA)$ is given by
\begin{align}
    \| f \|_2 = \left( \int_{\Omega'} |f|^2 w dA \right)^{1/2}
    \mbox{ for any $f \in L^2(wdA)$.}
    \label{0803.3}
\end{align}

\begin{theorem}\label{0517.A}
Let $0 < p < \infty$ and $N \ge1$.  For $\alpha, \beta > 0$, let $\Omega = [0,\alpha] \times [0,\beta]$.  A bounded function $f: \Omega \longrightarrow \mathbb{R}$ is continuous on $\Omega \setminus \{ (x,0): 0 \le x \le \alpha \}$ if and only if the induced persistence block map $\widetilde{f} : P_N(\Omega) \longrightarrow L^2(wdA)$ is continuous with respect to the p-Wasserstein distance $W_p$ on $P_N(\Omega)$ and the norm $\|\cdot\|_2$ on the Hilbert space $L^2(wdA)$.
\end{theorem}

\begin{proof}
Suppose the induced persistence block map $\widetilde{f} : P_N(\Omega) \longrightarrow L^2(wdA)$ is continuous.  To show the underlying function $f: \Omega \longrightarrow \mathbb{R}$ is continuous on $\Omega \setminus \{ (x,0): 0 \le x \le \alpha \}$, let $(a,b) \in \Omega$ with $b > 0$ and let $\epsilon >0$.  We are to show there is a   $\delta > 0$ such that 
\begin{align}
|f(x,y) - f(a,b)| < \epsilon, \mbox{ \ whenever $(x,y) \in \Omega$ with $\| (x,y) - (a,b) \|_{\infty} < \delta$.}
\nonumber 
\end{align}
To this end, by the continuity of the length function $\lambda$ and the assumption that $b > 0$, there exists a real number $\delta_1$ with $0 < \delta_1 < \lambda(a,b)/4$ such that
\begin{align}
\mbox{Area}(E(x,y) \cap E(a,b)) \ge \frac{1}{2}\mbox{Area}(E(a,b)) > 0, \ \text{whenever $ \| (x,y) - (a,b) \|_{\infty} < \delta_1$. }
\label{0517.A.1.1}
\end{align}
 Consider the compact square $K$ centered at $(a,b)$ given by 
\begin{align}
K = \{ (x,y) \in \Omega : \| (x,y) - (a,b) \|_{\infty} \le \delta_1 \}.
\nonumber
\end{align}
Since $0 < \delta_1 < \lambda(a,b)/4$, the compact square $K$ lies entirely above the $x$-axis, and hence $0 < \lambda(x,y) < 2y$ for all $(x, y)\in K$, by our definition of the length function $\lambda$.  
Furthermore, since $\lambda$ is continuous, the function $(x,y) \mapsto y - \frac{\lambda(x,y)}{2}$ is continuous and stays positive on the compact set $K$.  Thus, the set $\bigcup_{(x, y)\in K} E(x,y)$ is contained in a compact set that lies entirely above the $x$-axis.   Thus, by the continuity the functions $w$ and $\lambda$, there is a  real number $\eta > 0$ such that
\begin{align}
\eta & < \min \{ \lambda(x,y) : (x,y) \in K \}, \mbox{ and } \label{0517.A.2} \\
\eta & < \min \left\{ w(u, v) : (u, v) \in \bigcup_{(x, y)\in K} E(x,y) \right\}. \label{0517.A.3}
\end{align}

By assumption, the induced persistence block map $\widetilde{f} : P_N(\Omega) \longrightarrow L^2(wdA)$ is continuous at the persistence diagram $D_{0} = \{ (a,b) \}$  in $P_N(\Omega)$ that  consists of the single point $(a,b)$.  Thus, there exists a $\delta_2 > 0$ such that
\begin{align}
\| \widetilde{f}(D) - \widetilde{f}(D_0) \|_2 < \frac{ \eta^{3/2} \epsilon}{\sqrt{2}}, \ \text{whenever $W_p(D, D_0) < \delta_2$.}
\label{0517.A.4}
\end{align}
Select a $\delta > 0$ such that 
$0 < \max \{ \delta, \delta^p \} < \min \{ \delta_1, \delta_2 \}$.   
For any $(x,y) \in \Omega$ with 
\begin{align}
\| (x,y) - (a,b) \|_{\infty} < \delta,
\label{0517.A.5}
\end{align}
consider the persistence diagram $D = \{ (x,y) \} \in P_N(\Omega)$.  Observe that
\begin{align}
W_p (D, D_0) 
& \le
\begin{cases} 
\| (x,y) - (a,b) \|_{\infty}, & \text{ if $1 \le p < \infty$} \\
\| (x,y) - (a,b) \|_{\infty}^p, & \text{ if $0 < p < 1$}
\end{cases}
\nonumber \\
& < \max\{ \delta, \delta^p \} 
\nonumber \\
& < \delta_2.
\nonumber
\end{align}
Hence, the upper estimation for $\| \widetilde{f}(D) - \widetilde{f}(D_0) \|_2$ in (\ref{0517.A.4}) holds for any $(x,y) \in \Omega$ satisfying (\ref{0517.A.5}).
For a lower estimation, observe that
\begin{align}
\| \widetilde{f}(D) - \widetilde{f}(D_0) \|_2^2
& = \int_{\Omega'} \left| f(x,y) \chi_{E(x,y)} - f(a,b) \chi_{E(a,b)}\right|^2 w dA
\label{0517.A.7} \\
& \ge  \int_{E(x,y)\cap E(a,b)} \left| f(x,y)  - f(a,b) \right|^2 w dA
\nonumber \\
& \ge  \eta \mbox{Area}(E(x,y)\cap E(a,b)) \left| f(x,y)  - f(a,b) \right|^2, \mbox{ by (\ref{0517.A.3})}
\nonumber \\
& \ge \frac{ \eta}{2} \mbox{Area}(E(a,b))\left| f(x,y)  - f(a,b) \right|^2, \mbox{ by (\ref{0517.A.1.1}) }
\nonumber \\
& \ge \frac{ \eta^3}{2} \left| f(x,y)  - f(a,b) \right|^2, \mbox{ by (\ref{0517.A.2}). } 
\nonumber
\end{align}
Combining estimations in (\ref{0517.A.4}) and (\ref{0517.A.7}) yields
\begin{align}
\left| f(x,y)  - f(a,b) \right|
& \le \frac{\sqrt{2}}{ \eta^{3/2}} \| \widetilde{f}(D) - \widetilde{f}(D_0) \|_2
< \frac{\sqrt{2}}{ \eta^{3/2}} \cdot \frac{\epsilon  \eta^{3/2}}{\sqrt{2}} = \epsilon,
\nonumber
\end{align}
which completes the proof of the continuity of the function $f$.

Conversely, suppose the bounded function $f: \Omega \longrightarrow \mathbb{R}$ is continuous on $\Omega \setminus \{ (x,0) : 0 \le x \le \alpha \}$.  To prove the induced persistence block map $\widetilde{f} : P_N(\Omega) \longrightarrow L^2(wdA)$ is continuous, select a persistence diagram $D_0 \in P_N(\Omega)$ and an $\epsilon > 0$.  We are to show there is a real number $\delta > 0$ such that
\begin{align}
\| \widetilde{f}(D) - \widetilde{f}(D_0) \|_2 < \epsilon, \mbox{ \ whenever $D \in P_N(\Omega)$ with $W_p(D, D_0) < \delta$.}
\label{0517.A.8}
\end{align}

To this end, since the function $f$ is bounded and the functions $\lambda, w$ are continuous on the compact sets $\Omega = [0, \alpha] \times [0,\beta]$ and $\Omega' = [-\beta, \alpha + \beta] \times [0, 2\beta]$ respectively, we may select a constant $M > 0$ such that
\begin{align}
\| f \|_{\infty} & = \sup \{ |f(x,y)| : (x,y) \in \Omega \}  \le M, 
\label{0517.A.10} \\
\| \lambda \|_{\infty} & = \sup \{ \lambda(x,y) : (x,y) \in \Omega \}  \le M, 
\label{0517.A.11} \\
\| w \|_{\infty} & = \sup \left\{ w(x,y) : (x,y) \in \Omega' \right\}  \le M.
\label{0517.A.12}
\end{align}
Furthermore, by the continuity of the length function $\lambda$  on the compact set $\Omega$, the  function $\lambda$ is uniformly continuous on $\Omega$.  Thus, there exists a real number $\delta_1 > 0$ such that
\begin{align}
| \lambda(x,y) - \lambda(a,b) | < \frac{\epsilon^2}{192 M^{4} N^2}, \ \text{whenever $\| (x,y) - (a,b) \|_{\infty} < \delta_1$. }
\label{0517.A.12.1}
\end{align}
Next, by the definition of the length function in (\ref{kit.0618.1}), select a real number $c$ such that
\begin{align}
0 < c < \min \{ \lambda(a,b) : (a,b) \in D_0 \mbox{ with } b > 0 \} < 2\beta.
\label{0517.A.9}
\end{align}
 Likewise by the continuity on a compact set, the function $f$ is uniformly continuous on the compact set $ [0,\alpha] \times [\frac{c}{2}, \beta]$, and so there exists a real number $\delta_2 > 0$ such that  
\begin{align}
| f(x,y) - f(a,b) | < \frac{\epsilon^2}{192 M^{4} N^2}. 
\label{0517.A.13}
\end{align} 
whenever $(x,y), (a,b) \in [0,\alpha] \times [\frac{c}{2}, \beta]$ with $\| (x,y) - (a,b) \|_{\infty} < \delta_2$.
 Select a real number $\delta > 0$ such that
\begin{align}
0 < \max \{ \delta, \delta^{1/p} \} < \min \left\{ \frac{c}{2},\,  \delta_1,\,  \delta_2,\,  \frac{\epsilon^2}{192 M^4 N^2} \right\}.
\label{0517.A.14}
\end{align}

To show this is the desired $\delta$ for (\ref{0517.A.8}) to hold, let $D$ be any persistence diagram in $P_N(\Omega)$ with
\begin{align}
W_p(D, D_0) < \delta.
\label{0517.A.15}
\end{align}
Since $D$, $D_0$ are in $P_N(\Omega)$, the $p$-Wasserstein distance $W_p(D, D_0)$ is attained for a specific bijection $\gamma$.  Furthermore, in order to have a bijection $\gamma$ between $D$ and $D_0$, we may need to add at most $N$ points along the $x$-axis to each persistence diagram.  Thus, we may assume the persistence diagrams $D$, $D_0$ each has $m$ points with $1 \le m \le 2N$, and without loss of generality, we can assume the points in $D$ and $D_0$ are arranged as
\begin{align}
D_0 = \{ (a_j, b_j) : 1 \le j \le m \} \mbox{ and }
D = \{ (x_j, y_j) : 1 \le j \le m \}
\label{0517.A.30}
\end{align}
so that we use (\ref{pWasserdef1}) and (\ref{pWasserdef2}) to rewrite (\ref{0517.A.15}) as
\begin{align}
W_p(D,D_0) & = 
\begin{cases}
\left( \sum_{j=1}^{m} \| (x_j, y_j) - (a_j, b_j) \|_{\infty}^p \right)^{1/p}, & \text{ if $1 \le p < \infty$} \\
 \sum_{j=1}^{m} \| (x_j, y_j) - (a_j, b_j) \|_{\infty}^p , & \text{ if $0 < p < 1$}
\end{cases} 
\label{0517.A.31}  \\
& < \delta.
\nonumber
\end{align}
Moreover, we can assume, by the symmetry of the roles of $D$ and $D_0$ in (\ref{0517.A.31}), there exists an integer $m_0$ with $1 \le m_0 \le m$ and
\begin{align}
b_j > 0 \mbox{ if and only if } 1 \le j \le m_0.
\nonumber
\end{align}

In order to show $\| \widetilde{f}(D) - \widetilde{f}(D_0) \|_2 < \epsilon$, we first establish the following claim.
\bigskip

\noindent
\textit{Claim.} For integers $j$ with $1 \le j \le m$, we have
\begin{align}
\int_{\Omega'} \left| f(x_j, y_j) \chi_{E(x_j, y_j)} - f(a_j, b_j) \chi_{E(a_j, b_j)} \right|^2 w dA
< \frac{\epsilon^2}{16N^2}.
\nonumber
\end{align}
\medskip

\noindent
\textit{Proof of Claim.} 
In the  case that $1 \le m_0 < m$, we see that if an integer $j$ satisfies $m_0 < j \le m$, then we have $b_j = 0$ and $\lambda(a_j, b_j) = 0$, which in turn implies $\mbox{Area}(E(x_j,y_j) \cap E(a_j,b_j)) = 0$.  Thus, if  $m_0 < j \le m$,  by Lemma \ref{0617.A}, Statement (2), we have 
\begin{align}
\int_{\Omega'} \left| f(x_j, y_j) \chi_{E(x_j, y_j)} - f(a_j, b_j) \chi_{E(a_j, b_j)} \right|^2 w dA
& \le 2M^4 \| (x_j,y_j) - (a_j, b_j) \|_{\infty}
\nonumber \\
& \le 2M^4 \max\{ \delta, \delta^{1/p} \}, \mbox{ by (\ref{0517.A.31})}
\nonumber \\
& < 2M^4 \cdot \frac{\epsilon^2}{192M^4N^2},  \ \text{by (\ref{0517.A.14})} 
\nonumber \\
& < \frac{\epsilon^2}{16N^2},
\nonumber
\end{align}
which establishes the claim for the integers $j$ satisfying $m_0 <  j < m$.

For an integer $j$ with $1 \le j \le m_0$, we have $b_j > 0$ and by (\ref{0517.A.9}), we have $\lambda(a_j, b_j) \ge c > 0$.  It follows from (\ref{kit.0618.1}) that  $(a_j, b_j ) \in [0, \alpha] \times [\frac{c}{2}, \beta]$.  Also, observe that by (\ref{0517.A.31})
\begin{align}
\| (x_j, y_j) - (a_j, b_j) \|_{\infty} 
& \le
\begin{cases} 
W_p(D, D_0), & \text{ if $1 \le p < \infty$} \\
W_p(D, D_0)^{1/p}, & \text{ if $0 < p < 1$}
\end{cases}
\nonumber \\
& < \max \{ \delta, \delta^{1/p} \}
\nonumber \\
&  \le \frac{c}{2}, \mbox{ by (\ref{0517.A.14})}
\nonumber \\
&  \le \frac{\lambda(a_j,b_j)}{2},   \mbox{ by (\ref{0517.A.9})}
\nonumber  \\
& \le \frac{\lambda(x_j,y_j) + \lambda(a_j,b_j)}{2}, \mbox{ because $\lambda(x_j,y_j) \ge 0$.}
\nonumber
\end{align}
Hence, we have
$(x_j, y_j), (a_j, b_j) \in [0, \alpha] \times [\frac{c}{2}, \beta]$ with 
$\| (x_j, y_j) - (a_j, b_j) \|_{\infty} < \max\{ \delta, \delta^{1/p} \} $.  Furthermore, by Lemma \ref{0511.A}, we have 
$\mbox{Area}(E(x_j,y_j) \cap E(a_j,b_j)) > 0$.

Returning to the integral in our claim, observe that Statement (1) of Lemma \ref{0617.A} gives us 
\begin{align}
& \ \int_{\Omega'} |f(x_j ,y_j )\chi_{E(x_j ,y_j)} - f(a_j,b_j) \chi_{E(a_j,b_j)}|^2 wdA
\label{0517.A.1.1.1} \\
\le & \  4M^4\bigg( |f(x_j,y_j) - f(a_j,b_j)| + |\lambda(x_j,y_j) - \lambda(a_j,b_j)| 
+ \| (x_j,y_j) - (a_j, b_j) \|_{\infty} \bigg)
\nonumber \\
< & \ 4M^4 \left( \frac{\epsilon^2}{192M^4N^2} + \frac{\epsilon^2}{192M^4N^2} + \frac{\epsilon^2}{192M^4N^2} \right),
\mbox{ by (\ref{0517.A.13}), (\ref{0517.A.12.1}), (\ref{0517.A.14})}
\nonumber \\
= &\frac{\epsilon^2}{16N^2},    
\nonumber 
\end{align}
which concludes the proof of our claim.  $\Box$

Using the Claim, we now finish the proof of the theorem by showing $\| \widetilde{f}(D) - \widetilde{f}(D_0) \|_2 < \epsilon$.  Observe that
\begin{align}
\| \widetilde{f}(D) - \widetilde{f}(D_0) \|_2
& =
\left\| \sum_{j=1}^{m} f(x_j, y_j) \chi_{E(x_j, y_j)} - \sum_{j=1}^{m} f(a_j, b_j) \chi_{E(a_j, b_j)} \right\|_{2}
\nonumber \\
&  \le \sum_{j=1}^{m} \| f(x_j, y_j) \chi_{E(x_j, y_j)} - f(a_j, b_j) \chi_{E(a_j, b_j)}\|_2
\nonumber \\
& = \sum_{j=1}^{m} \left( \int_{\Omega'} |f(x_j, y_j) \chi_{E(x_j, y_j)} - f(a_j, b_j) \chi_{E(a_j, b_j)}|^2 w dA \right)^{1/2}.
\nonumber
\end{align}
In the case that $m_0 < m$, we use our claim to get
\begin{align}
\| \widetilde{f}(D) - \widetilde{f}(D_0) \|_2
& <  \sum_{j=1}^{m_0} \frac{\epsilon}{4N} + \sum_{j=m_0+1}^{m} \frac{\epsilon}{4N}
\label{0517.A.33} \\
& = m_0 \frac{\epsilon}{4N} + (m - m_0) \frac{\epsilon}{4N}
\nonumber \\
& \le 2N  \frac{\epsilon}{4N} + 2N \frac{\epsilon}{4N}
\nonumber \\
& =  \epsilon.
\nonumber
\end{align}
Lastly, in the case that $m_0=m$, we do not have the second summand on the right-hand side of (\ref{0517.A.33}), but the same argument gives $\| \widetilde{f}(D) - \widetilde{f}(D_0) \|_2 < \epsilon$.
\end{proof}

A continuous function $f:  [0, \infty) \times [0,\infty) \longrightarrow \mathbb{R}$ is always bounded on the compact set 
$\Omega = [0, \alpha]\times [0, \beta]$ with $\alpha, \beta > 0$ and continuous on the set $\Omega \setminus \{ (x,0) : 0 \le x \le \alpha \}$. Thus, Theorem \ref{0517.A} yields the following corollary.

\begin{corollary}\label{0525.A}
Let $0 < p < \infty$ and let $f : [0, \infty) \times [0,\infty) \longrightarrow \mathbb{R}$ be a continuous function. For any integer $N \ge 1$ and  $\Omega = [0, \alpha]\times [0, \beta]$ with $\alpha, \beta > 0$, the induced persistence block map $\widetilde{f} : P_N(\Omega) \longrightarrow L^2(wdA)$ is continuous with respect to the $p$-Wasserstein distance $W_p$ on $P_N(\Omega)$ and  the norm $\|\cdot\|_2$ of $L^2(wdA)$.
\end{corollary}

The corollary provides a theoretical result on the continuity of the induced persistence block map. On a practical note, 
we infer from the corollary that the computation of the vector $\widetilde{f}(D)$ for any given persistence diagram $D$ is robust with respect to the noises that incur in $D$.


\subsection{Identification Map}\label{identification map} 

We continue the study of induced persistence block maps $\widetilde{f}: P_N(\Omega) \longrightarrow L^2(wdA)$ in the special case that the underlying function $f$ is simply the constant function $1$. In this situation, the induced persistence block map becomes what we call the \textit{identification map} $\rho : P_N(\Omega) \longrightarrow L^2(wdA)$ given by
\begin{align}
\rho(D) = \sum_{(a,b) \in D} \chi_{E(a,b)}.
\label{0526.1}
\end{align}
Since the constant function is continuous on $\Omega$, by Theorem \ref{0517.A}, the identification map $\rho: P_N(\Omega) \longrightarrow L^2(wdA)$ is continuous with respect to the p-Wasserstein distance $W_p$ for any $p$ with $0 < p < \infty$.  When the length function $\lambda$ is Lipschitz, the identification map $\rho$ is stable with respect to the $1/2$-Wasserstein distance.

While one may be more familiar with the $2$-Wasserstein distance, the $1/2$-Wasserstein distance also naturally arises due to the square root in the definition of the Hilbert space norm on $L^2(wdA)$. In some situations, the $1/2$-Wasserstein distance provides a tighter bound than the $2$-Wasserstein distance.  For an illustration, take two persistence diagrams $D = \{ (1,1), (20,5) \}$ and $D_0 = \{ (2, 2), (20,1) \}$.  Observe that
\begin{align}
W_{1/2}(D,D_0) & = \| (1,1) - (2, 2) \|_{\infty}^{1/2} + \| (20,5) - (20,1) \|_{\infty}^{1/2}
\nonumber \\
& = \sqrt{1} + \sqrt{4} = 3,
\nonumber \\
W_{2}(D,D_0) & = \left( \| (1,1) - (2, 2) \|_{\infty}^{2} + \| (20,5) - (20,1) \|_{\infty}^{2}\right)^{1/2}
\nonumber \\
& = \sqrt{1^2 + 4^2} = \sqrt{17} \approx 4.12,
\nonumber
\end{align}
and so $W_{1/2}(D, D_0) < W_2(D, D_0)$.  Using the norm notation $\| \cdot \|_{2}$ of $L^2(wdA)$ given by (\ref{0803.3}), we state the main theorem of this subsection.

\begin{theorem}\label{0526.A}
Let $N \ge 1$, and let $\Omega = [0, \alpha] \times [0, \beta]$ and $\Omega' = [-\beta, \alpha + \beta] \times [0, 2\beta]$ for $\alpha, \beta > 0$. Suppose the length function 
$\lambda: \Omega \longrightarrow [0, \infty)$ is Lipschitz with respect to the sup-norm on $\Omega$ and $w: \Omega' \longrightarrow [0,\infty)$ satisfying (\ref{0802.1}).  Then, the identification map $\rho: P_N(\Omega) \longrightarrow L^2(wdA)$ is stable with respect to the $1/2$-Wasserstein distance.  In particular, for any positive real number $M$ which satisfies 
\begin{align}
& |\lambda(x,y) - \lambda(a,b)| \le M \| (x,y) - (a,b) \|_{\infty}  \ \text{whenever $(x,y), (a,b) \in \Omega$, and } 
\label{0526.A.1}  \\
&\|  \lambda \|_\infty  \le M \ \text{and} \ \|w\|_\infty \le M, 
\label{0526.A.2}
\end{align}
we have
\begin{align}
\left( \int_{\Omega} |\rho(D) - \rho(D_0)|^2 w dA \right)^{1/2}
\le \| \rho(D) - \rho(D_0) \|_2 
\le 2M \sqrt{M+1}\ W_{1/2}(D, D_0)
\label{0526.A.3}
\end{align}
for any $D, D_0 \in P_N(\Omega)$.
\end{theorem}

\begin{proof}
The first inequality in (\ref{0526.A.3}) is obvious because $\Omega \subseteq \Omega'$. To establish the second inequality (\ref{0526.A.3}), let $D$, $D_0$ be any two persistence diagrams in $P_N(\Omega)$.  Following the same reasons as in the proof of Theorem \ref{0517.A}, the $1/2$-Wasserstein distance $W_{1/2}(D,D_0)$ is attained for a specific bijection $\gamma: D\longrightarrow D_0$.  For this particular bijection $\gamma$ to exist, we may need to add at most $N$ points along the $x$-axis to each persistence diagram.  Hence, we may assume the persistence diagrams $D$, $D_0$ each has $m$ points with $1 \le m \le 2N$, and without loss of generality, the points in $D$ and $D_0$ are arranged such that
\begin{align}
D = \{ (x_j, y_j) : 1 \le j \le m \} \mbox{ and } D_0 = \{ (a_j, b_j) : 1 \le j \le m \},
\nonumber 
\end{align}
and
\begin{align}
W_{1/2}(D, D_0) = \sum_{j=1}^{m} \| (x_j, y_j) - (a_j, b_j) \|_{\infty}^{1/2}.
\nonumber
\end{align}
In order to estimate $\| \rho(D) - \rho(D_0) \|_2$,  observe that
\begin{align}
 \| \rho(D) - \rho(D_0) \|_2
=  &\ \left\| \sum_{j=1}^{m} \chi_{E(x_j, y_j)} - \sum_{j=1}^{m} \chi_{E(a_j,b_j)} \right\|_{2}
\label{0526.A.8} \\
\le & \ \sum_{j=1}^{m} \| \chi_{E(x_j, y_j)} - \chi_{E(a_j, b_j)} \|_{2}
\nonumber \\
= & \ \sum_{j=1}^{m} \bigg( \int_{\Omega'} | \chi_{E(x_j, y_j)} - \chi_{E(a_j, b_j)}|^2 w dA \bigg)^{1/2}
\nonumber \\
= & \  \sum_{j=1}^{m} \bigg( \int_{E(x_j, y_j) \setminus E(a_j, b_j)} \, w dA + \int_{E(a_j, b_j)\setminus E(x_j, y_j)} \, w dA \bigg)^{1/2}
\nonumber \\
\le & \  \sum_{j=1}^{m} \bigg( M \mbox{Area}(E(x_j, y_j) \setminus E(a_j, b_j)) + M \mbox{Area}(E(a_j, b_j) \setminus E(x_j, y_j)) \bigg)^{1/2}
\nonumber \\
= & \ \sum_{j=1}^{m} \sqrt{M} \bigg( \mbox{Area}(E(x_j, y_j) \setminus E(a_j, b_j)) + \mbox{Area}(E(a_j, b_j) \setminus E(x_j, y_j)) \bigg)^{1/2}.
\nonumber
\end{align}

For each integer $j$ in the above summation, we either have $\mbox{Area}(E(x_j, y_j) \cap E(a_j, b_j)) > 0$ or 
$\mbox{Area}(E(x_j, y_j) \cap E(a_j, b_j)) = 0$.  For the case when $\mbox{Area}(E(x_j, y_j) \cap E(a_j, b_j)) > 0$, by inequality (\ref{0617.A.5}) in the proof of Lemma \ref{0617.A}, we have
\begin{align}
& \ \mbox{Area}(E(x_j, y_j) \setminus E(a_j, b_j)) + \mbox{Area}(E(a_j, b_j) \setminus E(x_j, y_j))
\label{0526.A.1.1.1} \\
\le & \ 4M \bigg( | \lambda(x_j,y_j) - \lambda(a_j,b_j) | + \| (x_j,y_j) - (a_j,b_j) \|_{\infty} \bigg)
\nonumber \\
\le & \ 4M \bigg( M \| (x_j,y_j) - (a_j,b_j) \|_{\infty}+ \| (x_j,y_j) - (a_j,b_j) \|_{\infty} \bigg), \mbox{ by (\ref{0526.A.1})}
\nonumber \\
= & \ 4M(M+1)  \| (x_j,y_j) - (a_j,b_j) \|_{\infty}.
\nonumber
\end{align}

For the case when $\mbox{Area}(E(x_j, y_j) \cap E(a_j, b_j)) = 0$, by inequality (\ref{0617.A.7}) in the proof of Lemma \ref{0617.A}, we have
\begin{align}
& \ \mbox{Area}(E(x_j, y_j) \setminus E(a_j, b_j)) + \mbox{Area}(E(a_j, b_j) \setminus E(x_j, y_j))
\label{0526.A.1.1.2} \\
\le & \ 2M \| (x_j,y_j) - (a_j,b_j) \|_{\infty}
\nonumber \\
\le &  \ 4M(M+1) \| (x_j,y_j) - (a_j,b_j) \|_{\infty}.
\nonumber
\end{align}

Returning to the estimation of $\| \rho(D) - \rho(D_0) \|_2$ in (\ref{0526.A.8}), the above inequalities 
(\ref{0526.A.1.1.1}) and (\ref{0526.A.1.1.2}) give us
\begin{align}
\| \rho(D) - \rho(D_0) \|_2
& \le \sum_{j=1}^{m} \sqrt{M} \bigg( \mbox{Area}(E(x_j, y_j) \setminus E(a_j, b_j)) + \mbox{Area}(E(a_j, b_j) \setminus E(x_j, y_j)) \bigg)^{1/2}
\nonumber \\
& \le \sum_{j=1}^{m} \sqrt{M} \bigg(  4M(M+1) \| (x_j,y_j) - (a_j,b_j) \|_{\infty} \bigg)^{1/2}
\nonumber \\
& = 2M \sqrt{M+1}\sum_{j=1}^{m} \| (x_j,y_j) - (a_j,b_j) \|_{\infty}^{1/2}
\nonumber \\
& = 2M \sqrt{M+1} \ W_{1/2}(D,D_0),
\nonumber
\end{align}
which finishes the whole proof of the Theorem.
\end{proof}

The identification map $\rho$ is defined on $P_N(\Omega)$, but the estimation in inequality (\ref{0526.A.3}) in Theorem \ref{0526.A} is independent of the integer $N$.  From this observation, we can say the identification map $\rho: P(\Omega) \longrightarrow L^2(wdA)$ is stable with respect to the $1/2$-Wasserstein distance. While we initially establish the stability of the identification map $\rho$ relative to the $1/2$-Wasserstein distance, we can extend our stability result using any $p$-Wasserstein distance for $\frac{1}{2} < p < \infty$ when we restrict to $P_N(\Omega)$ for any fixed integer $N \ge 1$.
To achieve the estimation, we require H\"older's Inequality:  for any vectors 
$x = (x_1, x_2, \dots, x_n)$ and $y = (y_1, y_2, \dots, y_n)$ in $\mathbb{R}^n$, we have
\begin{align}
\sum_{j=1}^{n} |x_j y_j | \le
\left( \sum_{j=1}^{n} |x_j|^{q_0} \right)^{1/q_0}\left( \sum_{j=1}^{n} |y_j|^{p_0} \right)^{1/p_0},
\label{0527.1}
\end{align}
where $p_0,q_0$ satisfy $1 < p_0,q_0 < \infty$ and $\frac{1}{p_0} + \frac{1}{q_0} = 1$.  Keeping H\"older's inequality in mind, we have the following corollary to Theorem \ref{0526.A}.

\begin{corollary}\label{0527.A}
Let $N \ge 1$, and let $\Omega = [0, \alpha] \times [0, \beta]$ and $\Omega' = [-\beta, \alpha + \beta] \times [0, 2\beta]$ for $\alpha, \beta > 0$. Suppose the length function 
$\lambda: \Omega \longrightarrow [0, \infty)$ and the weight function $w: \Omega' \longrightarrow (0, \infty)$ satisfy the hypotheses of Theorem \ref{0526.A}.  The identification map $\rho: P_N(\Omega) \longrightarrow L^2(wdA)$ satisfies
\begin{align}
\left( \int_{\Omega} |\rho(D) - \rho(D_0)|^2 w dA \right)^{1/2}
 \le \| \rho(D) - \rho(D_0) \|_2
\le
\begin{cases}
4N M \sqrt{M+1} \ W_{p}(D, D_0)^{1/2}, & \text{ if $1 \le p < \infty$} \\
4N M \sqrt{M+1} \ W_{p}(D, D_0)^{1/2p}, & \text{ if $\frac{1}{2} < p < 1$}
\nonumber
\end{cases},
\end{align}
for any $D, D_0 \in P_N(\Omega)$.  In particular for $p=2$, we have
\begin{align}
\left( \int_{\Omega} |\rho(D) - \rho(D_0)|^2 w dA \right)^{1/2}
\le \| \rho(D) - \rho(D_0) \|_2 \le 4N M \sqrt{M+1} \ W_{2}(D, D_0)^{1/2}.
\nonumber
\end{align}
\end{corollary}

\begin{proof}
Let the real number $p$ satisfy $\frac{1}{2} < p < \infty$.  As in the proof of Theorem \ref{0526.A}, for any two persistence diagrams $D, D_0 \in P_N(\Omega)$, select the bijection $\gamma: D \longrightarrow D_0$ and the integer $m$ with $1 \le m \le 2N$ so that we arrange the points in $D$ and $D_0$ as
\begin{align}
D = \{ (x_j, y_j) : 1 \le j \le m \} \mbox{ and } D_0 = \{ (a_j, b_j) : 1 \le j \le m \},
\nonumber 
\end{align}
for which
\begin{align}
W_{p}(D, D_0) =
\begin{cases} 
\left( \sum_{j=1}^{m} \| (x_j, y_j) - (a_j, b_j) \|_{\infty}^{p}\right)^{1/p}, & \text{ if $1 \le p < \infty$} \\
\sum_{j=1}^{m} \| (x_j, y_j) - (a_j, b_j) \|_{\infty}^{p}, & \text{ if $\frac{1}{2} < p < 1.$}
\end{cases}
\label{0527.A.2}
\end{align}

It follows from (\ref{0526.A.3}) that 
\begin{align}
\| \rho(D) - \rho(D_0) \|_2 
& \le 2M\sqrt{M+1} \sum_{j=1}^{m} \| (x_j, y_j) - (a_j, b_j) \|_{\infty}^{1/2}
\label{0527.A.3} \\
& =  \sum_{j=1}^{m} 2M\sqrt{M+1} \| (x_j, y_j) - (a_j, b_j) \|_{\infty}^{1/2}.
\nonumber 
\end{align}
Applying H\"older's inequality to (\ref{0527.A.3}) with $p_0 = 2p > 1$ and $q_0 = \frac{1}{1 - \frac{1}{p_0}} = \frac{2p}{2p-1} > 1$ yields
\begin{align}
& \ \| \rho(D) - \rho(D_0) \|_2
\nonumber \\
\le & \ \left( \sum_{j=1}^{m}(2M\sqrt{M+1})^{q_0} \right)^{1/q_0}
\left(  \sum_{j=1}^{m}\| (x_j, y_j) - (a_j, b_j) \|_{\infty}^{p_0/2} \right)^{1/p_0}
\nonumber \\
= & \ 2 m^{1/q_0} M \sqrt{M+1} 
\left(  \sum_{j=1}^{m}\| (x_j, y_j) - (a_j, b_j) \|_{\infty}^{p} \right)^{1/2p}, \mbox{ because $p_0=2p$}
\nonumber \\
\le & \  4N M \sqrt{M+1} \left(  \sum_{j=1}^{m}\| (x_j, y_j) - (a_j, b_j) \|_{\infty}^{p} \right)^{1/2p}, \ \mbox{ because $m^{1/q_0} \le (2N)^{1/q_0} \le 2N$}
\nonumber \\
= & 
\begin{cases}
4N M \sqrt{M+1} \ W_p(D, D_0)^{1/2}, & \text{ if $1 \le p < \infty$} \\
4N M \sqrt{M+1} \ W_p(D, D_0)^{1/2p}, & \text{ if $\frac{1}{2} < p < 1,$}
\end{cases} \ \mbox{ by (\ref{0527.A.2})}.
\nonumber 
\end{align}
\end{proof}

It follows from the above theorem and corollary that with respect to $W_p(D, D_0)$ for any $p \geq 1/2$, the computation of $\rho(D)$ for any persistence diagram $D$ is robust with respect to the noises that incur in $D$, under an assumption on the length function $\lambda$ and the weight function $w$.


\subsection{Vectorization}\label{vectorization}

Now, for a persistence diagram $D$, we would like to associate every persistence block $\widetilde{f}(D)$, which is some vector in the Hilbert space $L^2(wdA)$, with a vector in $\R^n$. There are two ways to do that, depending whether one wants to focus on
$\Omega = [0, \alpha] \times [0, \beta]$ or $\Omega' = [-\beta, \alpha + \beta] \times [0, 2\beta]$ as discussed in the previous two subsections.  Since both ways share the same mathematical theory, we develop them together as one in this subsection.  To this end, select a partition $\mathcal P$ of either $\Omega$ or $\Omega'$, whichever one chooses to focus, into $n$ grid cells $S_1, S_2, \ldots, S_n$. For any persistence diagram $D$ in $P(\Omega)$, let 
\begin{eqnarray}
I_{f, S_i} (D) = \int_{S_i} \, \widetilde{f}(D) \, wdA,
\nonumber
\end{eqnarray}
and define the vector 
\begin{eqnarray}\label{kit.0618.5}
\widetilde{f}_{\mathcal P}(D) =  (I_{f, S_1}(D), I_{f, S_2}(D), \ldots, I_{f, S_n}(D))
\nonumber
\end{eqnarray} 
in $\R^n$ to be the {\it vectorized persistence block} (VPB).  Thus we can think of $\widetilde{f}_{\mathcal P}$ as a map $\widetilde{f}_{\mathcal P}: P(\Omega) \longrightarrow \R^n$.

To ensure the theory accounts for both types of partitions, let
\begin{align}
    G =
    \begin{cases}
    \Omega, & \text{ if $\mathcal{P}$ is a partition of $\Omega$} \\
    \bigcup_{(x,y) \in \Omega} E(x,y), & \text{ if $\mathcal{P}$ is a partition of $\Omega'$}
    \end{cases}.
    \label{0803.1}
\end{align}
Note that $\Omega \subseteq G \subseteq \Omega'$, and the persistence block $\widetilde{f}(D)$ is a linear combination of characteristic functions $\chi_{E(x,y)}$ of the square $E(x,y)$ centered at $(x,y) \in \Omega$.  Thus, whether $S_1, S_2, \dots, S_n$ form a partition $\mathcal{P}$ of $\Omega$ or $\Omega'$, we have
\begin{align}
    I_{f, S_i}(D) & = \int_{S_i} \widetilde{f}(D) wdA 
    \label{0803.2} \\
    & = \int_{S_i \cap G} \widetilde{f}(D) wdA.
    \nonumber 
\end{align}

The VPB $\widetilde{f}_{\mathcal P}(D)$ is a vector in $\mathbb{R}^n$, and so we may use the standard norms on $\mathbb{R}^n$ to measure the distance between two VPBs.  In particular, for any two persistence diagrams $D, D_0 \in P_N(\Omega)$ and  for $1 \le p < \infty$, we have 
\begin{align}
\| \widetilde{f}_{\mathcal P}(D) - \widetilde{f}_{\mathcal P}(D_0) \|_p 
& = \left( \sum_{i=1}^{n} | I_{f, S_i}(D) - I_{f, S_i}(D_0)|^p \right)^{1/p},
\label{0615.1}
\end{align}
and
\begin{align}
\| \widetilde{f}_{\mathcal P}(D) - \widetilde{f}_{\mathcal P}(D_0) \|_{\infty} 
& =  \sup_{1 \le i \le n} | I_{f, S_i}(D) - I_{f, S_i}(D_0)|.
\label{0615.2}
\end{align}

The induced persistence block map $\widetilde{f}: P_N(\Omega) \longrightarrow L^2(wdA)$ is continuous whenever the underlying function $f: [0,\infty) \times [0,\infty) \longrightarrow \R$ is bounded and continuous, by Theorem \ref{0517.A}. 
In order to explore a similar continuity result for the VPB relative to the Hilbert space norm $\| \cdot \|_2$ given in (\ref{0615.1}) above, we establish an upper estimate for $\|  \widetilde{f}_{\mathcal P}(D) -  \widetilde{f}_{\mathcal P}(D_0) \|_2$, which involves the measure $\mu$ defined by
\begin{align}
\mu(E) = \int_E d\mu = \int_E wdA.
\nonumber
\end{align}

\begin{lemma}\label{0615.A}
Let $\Omega = [0, \alpha] \times [0, \beta]$ and $\Omega' = [-\beta, \alpha+\beta] \times [0, 2 \beta]$ with $\alpha, \beta > 0$, let $\mathcal{P}$ be a partition of \ $\Omega$ or $\Omega'$ into $n$ grid cells $S_1, S_2, \dots, S_n$, and let $G$ be the set defined in (\ref{0803.1}).  For persistence diagrams $D = \{ (x_j, y_j) : 1 \le j \le m \}$ and $D_0 = \{ (a_j, b_j) : 1 \le j \le m \}$, we have
\begin{align}
\|  \widetilde{f}_{\mathcal P}(D) -  \widetilde{f}_{\mathcal P}(D_0) \|_2^2
& \le \sum_{i=1}^{n} \mu(S_i\cap G) \left[ \sum_{j=1}^{m} \left( 
\int_{S_i} | f(x_j, y_j)\chi_{E(x_j,y_j)} - f(a_j, b_j) \chi_{E(a_j, b_j)}|^2 wdA
\right)^{1/2} \right]^2.
\nonumber
\end{align}
\end{lemma}

\begin{proof}
Observe that for integers $i$ with $1 \le i \le n$, we have
\begin{align}
&\ \left| I_{f, S_i}(D) - I_{f, S_i}(D_0) \right|^2
\nonumber \\
= &\ \left| \int_{S_i\cap G} \widetilde{f}(D)wdA - \int_{S_i\cap G} \widetilde{f}(D_0)wdA \right|^2, \mbox{ by (\ref{0803.2})}
\nonumber  \\
\le & \ \left( \int_{S_i\cap G} |\widetilde{f}(D) - \widetilde{f}(D_0)|wdA \right)^2
\nonumber \\
\le &\  \left( \int_{S_i \cap G} 1^2 wdA \right) \left( \int_{S_i\cap G} |\widetilde{f}(D) - \widetilde{f}(D_0)|^2wdA \right), 
\mbox{ by the Cauchy-Schwarz Inequality}
\nonumber \\
\le & \ \mu(S_i \cap G) \int_{S_i} |\widetilde{f}(D) - \widetilde{f}(D_0)|^2wdA,
\mbox{ because $S_i \cap G \subseteq S_i$ and the definition of $\mu$}
\nonumber \\
\le & \ \mu(S_i \cap G) \left( \sum_{j=1}^{m} \| (f(x_j,y_j)\chi_{E(x_j,y_j)} - f(a_j, b_j)\chi_{E(a_j,b_j)})\chi_{S_i} \|_2 \right)^2, \ \text{by the Triangle Inequality}
\nonumber \\
= & \ \mu(S_i \cap G) \left( \sum_{j=1}^{m} \left( \int_{S_i} | (f(x_j,y_j)\chi_{E(x_j,y_j)} - f(a_j, b_j)\chi_{E(a_j,b_j)}|^2 wdA \right)^{1/2}\right)^2 .
\nonumber 
\end{align}
Summing over all the grid cells $S_i$ in the partition $\mathcal{P}$ leads to the desired inequality.
\end{proof}

Similar to the identification map $\rho$, under some basic assumptions, we now show the vectorized persistence block map $\widetilde{f}_{\mathcal P}: P(\Omega) \longrightarrow \mathbb{R}^n$ is stable relative to the $1/2$-Wasserstein distance.

\begin{theorem}\label{0615.B}
Let $\Omega = [0, \alpha] \times [0, \beta]$ and $\Omega' = [-\beta, \alpha+\beta] \times [0, 2 \beta]$ with $\alpha, \beta > 0$, let $\mathcal{P}$ be a partition of \ $\Omega$ or $\Omega'$ into $n$ grid cells $S_1, S_2, \dots, S_n$, and let $G$ be the set defined in (\ref{0803.1}).  Suppose the function $f: \Omega \longrightarrow \mathbb{R}$ and the length function $\lambda: \Omega \longrightarrow [0, \infty)$ are Lipschitz.  Then the vectorized persistence block map $\widetilde{f}_{\mathcal{P}}: P(\Omega) \longrightarrow \mathbb{R}^n$ is stable relative to the $1/2$-Wasserstein distance.  In particular, for any positive real number $M$ which satisfies
\begin{align}
& |f(x,y) - f(a,b)| \le M \| (x,y) - (a,b) \|_{\infty} \ \text{for all $(x,y), (a,b) \in \Omega$; }
\label{0615.B.1} \\
& |\lambda(x,y) - \lambda(a,b)| \le M \| (x,y) - (a,b) \|_{\infty} \ \text{for all $(x,y), (a,b) \in \Omega$; }
\label{0615.B.2} \\
& \|f \|_\infty \le M, \  \|  \lambda\|_\infty  \le M , \  \text{and} \  \|w\|_\infty  \le M, 
\label{0615.B.3}
\end{align}
we have
\begin{align}
\| \widetilde{f}_{\mathcal{P}}(D) - \widetilde{f}_{\mathcal{P}}(D_0) \|_2 \le 2M^2 \sqrt{\mu(G)(2M+1)}\ W_{1/2}(D, D_0),
\label{0615.B.4}
\end{align}
for any $D, D_0 \in P(\Omega)$.
\end{theorem}

\begin{proof}
Assume the positive real number $M$ satisfies inequalities (\ref{0615.B.1}), (\ref{0615.B.2}) and (\ref{0615.B.3}).  For any two persistence diagrams $D, D_0 \in P(\Omega)$, let $N$ be the maximum cardinality of $D$ and $D_0$.  As in the proof of previous results, we may assume $D,D_0$ each has $m$ points with $1 \le m \le 2N$, and the points in $D,D_0$ are arranged such that
\begin{align}
D = \{ (x_j, y_j) : 1 \le j \le m \} \mbox{ and } D_0 = \{ (a_j, b_j) : 1 \le j \le m \}
\nonumber
\end{align}
and 
\begin{align}
W_{1/2}(D,D_0) = \sum_{j=1}^{m} \| (x_j, y_j) - (a_j, b_j) \|_{\infty}^{1/2}.
\nonumber
\end{align}


As in the proof of Theorem \ref{0526.A}, we have two cases based on the area of 
$E(x_j, y_j) \cap E(a_j, b_j)$.  For the case when 
$\mbox{Area}(E(x_j, y_j) \cap E(a_j, b_j)) > 0$, Lemma \ref{0617.A} and assumptions (\ref{0615.B.1}), (\ref{0615.B.2}) give us
\begin{align}
& \ \int_{S_i} | f(x_j, y_j)\chi_{E(x_j,y_j)} - f(a_j, b_j) \chi_{E(a_j, b_j)}|^2 wdA
\nonumber \\
\le & \ \int_{\Omega'} | f(x_j, y_j)\chi_{E(x_j,y_j)} - f(a_j, b_j) \chi_{E(a_j, b_j)}|^2 wdA,
\mbox{ because $S_i \subseteq \Omega'$}
\nonumber \\
\le & \ 4M^4\bigg( |f(x_j,y_j) - f(a_j,b_j)| + |\lambda(x_j,y_j) - \lambda(a_j,b_j)| 
+ \| (x_j,y_j) - (a_j, b_j) \|_{\infty} \bigg)
\nonumber \\
\le & \ 4M^4\bigg( M \| (x_j,y_j) - (a_j, b_j) \|_{\infty} + M \| (x_j,y_j) - (a_j, b_j) \|_{\infty}
+ \| (x_j,y_j) - (a_j, b_j) \|_{\infty} \bigg)
\nonumber \\
= & \ 4M^4(2M+1) \| (x_j,y_j) - (a_j, b_j) \|_{\infty}.
\nonumber 
\end{align}

For the case when $\mbox{Area}(E(x_j, y_j) \cap E(a_j, b_j)) = 0$, Lemma \ref{0617.A} gives us
\begin{align}
& \ \int_{S_i} | f(x_j, y_j)\chi_{E(x_j,y_j)} - f(a_j, b_j) \chi_{E(a_j, b_j)}|^2 wdA
\nonumber \\
\le & \ \int_{\Omega'} | f(x_j, y_j)\chi_{E(x_j,y_j)} - f(a_j, b_j) \chi_{E(a_j, b_j)}|^2 wdA
\nonumber \\
\le & \ 2M^4\ \| (x_j,y_j) - (a_j, b_j) \|_{\infty}
\nonumber \\
\le & \ 4M^4(2M+1) \| (x_j,y_j) - (a_j, b_j) \|_{\infty}.
\nonumber 
\end{align}

To finish the estimation of 
$\| \widetilde{f}_{\mathcal{P}}(D) - \widetilde{f}_{\mathcal{P}}(D_0) \|_2$ 
using Lemma \ref{0615.A}, observe that
\begin{align}
\| \widetilde{f}_{\mathcal{P}}(D) - \widetilde{f}_{\mathcal{P}}(D_0) \|_2^2
& \le \sum_{i=1}^{n} \mu(S_i \cap G) \left[ \sum_{j=1}^{m} \left( 
\int_{S_i} | f(x_j, y_j)\chi_{E(x_j,y_j)} - f(a_j, b_j) \chi_{E(a_j, b_j)}|^2 wdA \right)^{1/2} \right]^2
\nonumber \\
& \le \sum_{i=1}^{n} \mu(S_i \cap G) \left[ \sum_{j=1}^{m} \left( 
4M^4(2M+1) \| (x_j,y_j) - (a_j,b_j) \|_{\infty} \right)^{1/2} \right]^2
\nonumber \\
& = 4M^4(2M+1) \sum_{i=1}^{n} \mu(S_i \cap G) 
\left[ \sum_{j=1}^{m} \| (x_j,y_j) - (a_j,b_j) \|_{\infty}^{1/2}  \right]^2
\nonumber \\
& = 4M^4(2M+1) \ W_{1/2}(D, D_0)^2 \sum_{i=1}^{n} \mu(S_i \cap G)
\nonumber \\
& = 4M^4\mu(G)(2M+1)\ W_{1/2}(D, D_0)^2,
\nonumber
\end{align}
which  concludes the proof.
\end{proof}

Similar to the identification map $\rho$ discussed in the previous subsection, the stability of the vectorized persistence block map $\widetilde{f}_{\mathcal{P}}$ relative to the $1/2$-Wasserstein distance is independent of the number of points in the persistence diagrams $D, D_0 \in P(\Omega)$.  If we restrict our attention to the persistence diagrams in $P_N(\Omega)$, the following corollary shows the vectorized persistence block map $\widetilde{f}_{\mathcal{P}}$ is stable relative to the $p$-Wasserstein distance for $\frac{1}{2} < p < \infty$. The proof of the corollary utilizes H\"older's inequality, following the exact same argument as in the proof of Corollary \ref{0527.A}, and we need not repeat the argument here.

\begin{corollary}\label{0615.C}
Assuming the hypotheses of Theorem \ref{0615.B}, the VPBs $\widetilde{f}_{\mathcal{P}}(D)$ and $\widetilde{f}_{\mathcal{P}}(D_0)$ satisfy
\begin{align}
\| \widetilde{f}_{\mathcal{P}}(D) - \widetilde{f}_{\mathcal{P}}(D_0) \|_2
\le
\begin{cases}
4N M^2 \sqrt{\mu(G)(2M+1)} \ W_{p}(D, D_0)^{1/2}, & \text{ if $1 \le p < \infty$} \\
4N M^2 \sqrt{\mu(G)(2M+1)} \ W_{p}(D, D_0)^{1/2p}, & \text{ if $\frac{1}{2} < p < 1$}
\end{cases},
\nonumber 
\end{align}
for any $D, D_0 \in P_N(\Omega)$.  In particular for $p=2$, we have
\begin{align}
\| \widetilde{f}_{\mathcal{P}}(D) - \widetilde{f}_{\mathcal{P}}(D_0) \|_2 
\le 4 N M^2 \sqrt{\mu(G)(2M+1)} \ W_{2}(D, D_0)^{1/2}.
\nonumber
\end{align}
\end{corollary}

The vectorization map $\widetilde{f}_\mathcal P (D)$ is stable with respect to the $p-$Wasserstein distance, when $1/2 \leq p < \infty$. The estimations provided in Theorem \ref{0615.B} and Corollary \ref{0615.C}, do not even depend on the number $n$ of grid cells in the partition $\mathcal P$ of $\Omega$ or $\Omega'$.  In practice, we can take $\mu(G)$ to be a fixed constant and we can have a maximum number $N$ of 
points in a persistence diagram.   Thus there is a natural upper bound for the integer $N$ in our estimations in the conclusion of Corollary \ref{0615.C}.  However,  in the case for $p = 1/2$, the estimation 
provided in Theorem \ref{0615.B} does not depend on the integer $N$.

%% file: Experiments-2.tex

In this section we present our simulation results to illustrate the utility of vectorized persistence blocks (VPB) for classification, clustering and change point detection tasks. In all of the simulations, our results are compared with the corresponding results produced from persistence images (PI). Two of the simulations closely mimic those of Sections 6.1 and 6.4 in \cite{PI}. The simulations involve both synthetic and real datasets. On 2D datasets, we extract topological information using the Vietoris-Rips filtration \cite{ghrist2008barcodes}. For large 3D datasets, we resort to the $\alpha$-shape complexes \cite{edelsbrunner1994three,edelsbrunner2010computational} to avoid computational challenges that arise from using the Vietoris-Rips complexes in that situation. 

To compute VPBs, we experiment with several choices of the weight function $w$ and the side length function $\lambda$. For example, for $\lambda$ we consider functions of the form  
\[\lambda=F(y; \tau,n,m) =2 \tau y \left(\frac{y}{y_{max}}\right)^{n} \left(1-\frac{y}{y_{max}}\right)^{m},\]
where $\tau\in(0,1]$, $n$, $m$ are non-negative integers and $y_{max}$ is the maximum of  persistence values $y$ for the PD in question. All our results are based on the weight function $w(x,y)=x+y$ and the side length $\lambda=F(y;\tau,0,0)=2\tau y$, where $0<\tau\leq1$. Note that with the selected functions for weight and side length, VPBs can be computed analytically. PDs are computed using the {\tt R} packages {\tt TDA} \cite{fasy2021package} and {\tt TDAstats} \cite{wadhwa2018tdastats}. To produce PIs we utilize the {\tt R} package {\tt kernelTDA} \cite{kernelTDApackage}. 

\subsection{Clustering using K-medoids on a synthetic dataset}\label{sec1.1}
This simulation study considers the synthetic dataset used in Sections 6.1 of \cite{PI} which consists of 25 point clouds of size 500 sampled uniformly from each of the following six shapes: a unit cube, a circle of radius 1/2, a sphere of radius 1/2, three clusters with random centers in the unit cube, three clusters each containing another set of three clusters, a torus with inner and outer diameters of 1/2 and 1 respectively. Furthermore, Gaussian noise is added to each point cloud at two levels: $\eta=0.05$ and $\eta=0.1$. Topological signatures (in the form of VPBs and PIs) are computed from PDs for homological dimensions $H_0$ and $H_1$. In this study, the goal is to cluster the given 150 point clouds into 6 respective shape classes. 

For PIs, we use the parameter settings adopted by the authors of \cite{PI}: standard deviation $\sigma=0.1$, a $20\times20$ grid and the linear weighting function. VPBs are computed with the candidate values of $\tau$ ranging in $\{0.1,0.3,0.5,0.7,0.9\}$ on a grid of size $6\times 6$. The final value of $\tau$ is selected based on the internal measure of the Davies-Bouldin Index \cite{halkidi2002clustering}. More specifically, among the five different clustering performances (one for each $\tau$), we choose that for which the Davies-Bouldin Index is minimum.

Following the approach in \cite{PI}, clustering is performed employing K-medoids method \cite{rdusseeun1987clustering,park2009simple}, which is a robust version of the well-known K-means algorithm \cite{lloyd1982least}. It uses actual data points as cluster centers (called medoids) instead of
centroids as in K-means and can be applied with any dissimilarity measure. To implement K-medoids in {\tt R}, we use {\tt cluster::pam()} function with its default parameter settings \cite{clusterRpackage}.

A $150 \times 150$ matrix of pairwise distances (also called a dissimilarity matrix) between topological signatures (VPBs or PIs) is computed and passed to K-medoids (specifying the number of clusters to be six) which returns a list of cluster assignments. The performance is evaluated based on the accuracy of cluster predictions against the true cluster labels. 
%

\begin{figure}
      \centering
    \begin{minipage}[b]{.45\textwidth}
  \centering
  \begin{tabular}{ccccc}
    \hline
    Metric & Dim & Noise & PI & VPB \\ 
    \hline
    $L_1$ & $H_0$ & 0.05 & 0.933 &  \bf{0.947} \\ 
    $L_2$ & $H_0$ & 0.05 & 0.927 &   \bf{0.940} \\ 
    $L_\infty$ & $H_0$ & 0.05 & 0.940   & \bf{0.947} \\ 
    $L_1$ & $H_0$ & 0.10 & 0.953 &  \bf{0.960}\\ 
    $L_2$ & $H_0$ & 0.10 & 0.953 &  \bf{0.960} \\ 
    $L_\infty$ & $H_0$ & 0.10 & \bf{0.960}  &  \bf{0.960}  \\ 
    $L_1$ & $H_1$ & 0.05 & \bf{1.000} & \bf{1.000} \\ 
    $L_2$ & $H_1$ & 0.05 & \bf{1.000} & \bf{1.000} \\ 
    $L_\infty$ & $H_1$ & 0.05 & \bf{1.000} &  \bf{1.000} \\ 
    $L_1$ & $H_1$ & 0.10 & 0.960 & \bf{0.973} \\ 
    $L_2$ & $H_1$ & 0.10 & 0.960 & \bf{0.973}\\ 
    $L_\infty$ & $H_1$ & 0.10 & \bf{0.960} &  \bf{0.960} \\ 
    \hline
  \end{tabular}
        \caption{K-medoids accuracy results.}
        \label{six-shapes-table}
      \end{minipage}%
      \quad
    \begin{minipage}[b]{.45\textwidth}
        \centering
        \includegraphics[scale=0.5]{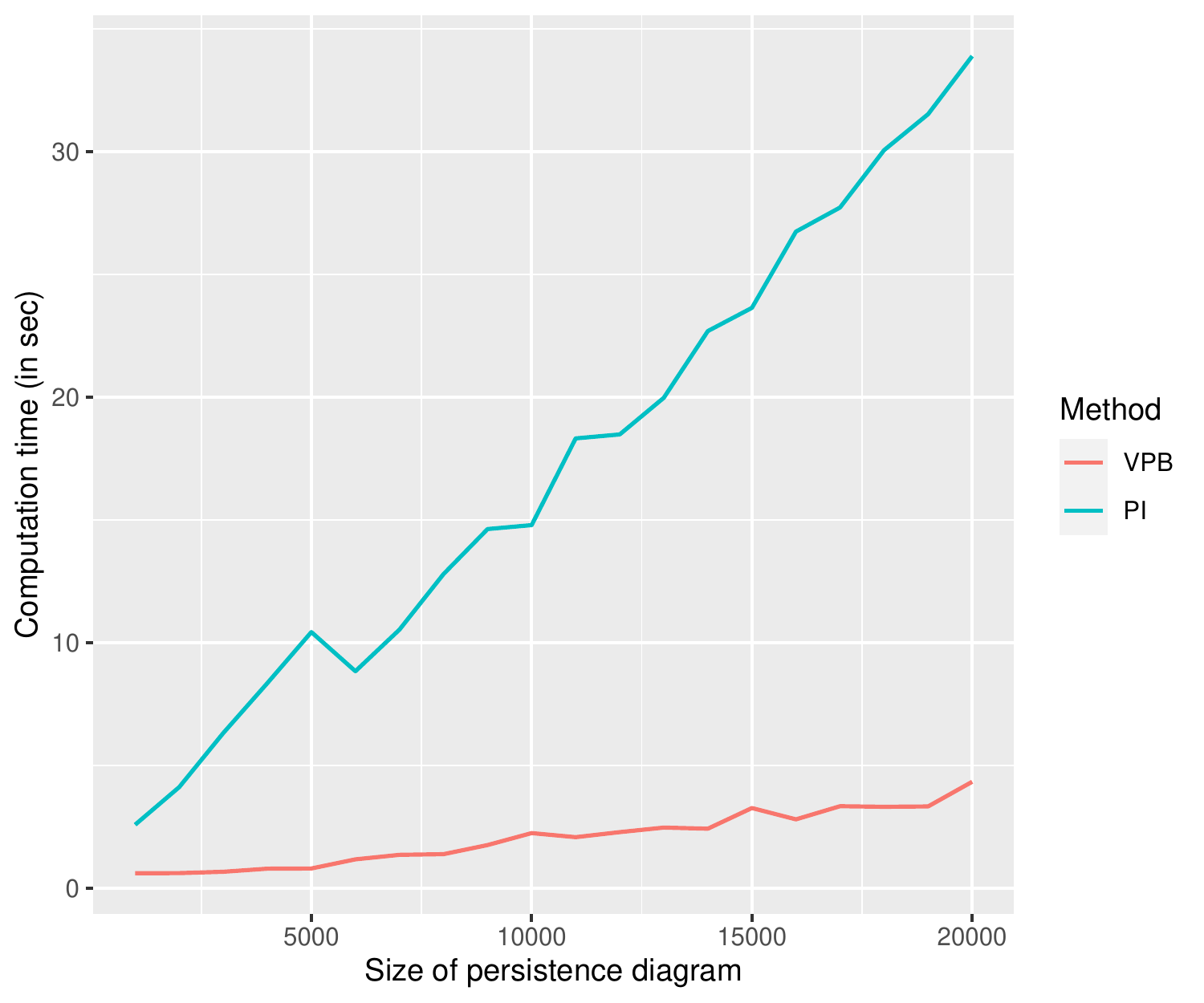}
        \caption{Comparison of computational cost.}
        \label{vpb_vs_pi_cost}
    \end{minipage}
\end{figure}

Comparison of the clustering accuracies is given in Figure \ref{six-shapes-table}. The results show that K-medoids produces slightly better accuracies on VPBs. Also, we observe that the computational cost associated with VPBs (0.09 and 0.188 seconds for $H_0$ and $H_1$) is lower than that of PIs (0.615 and 0.714 seconds for $H_0$ and $H_1$) despite running K-medoids five times for different values of $\tau$. Here, the cost is the combined time to compute the topological signatures and to run the K-medoids algorithm. Here, however, the observed difference in computational cost is partially due to the grid sizes being unequal (i.e. $6\times 6$ for VPBs and $20\times 20$ for PIs). So next, we compare the computational times to produce VPBs and PIs computed over the same grid. We generate 100 PDs of size $n\in\{1000,2000,\ldots,19000,20000\}$, where the birth and persistence values are randomly sampled from the beta distributions with the shape parameters $\alpha_b=4$, $\beta_b=6$ and $\alpha_p=1$, $\beta_p=5$ respectively. We superimpose a grid of size $10 \times 10$ to compute VPS and PIs. The plot of the total computational cost (in seconds) versus the size of PDs is given in Figure \ref{vpb_vs_pi_cost}, which shows that VBPs are 6-7 times faster to compute than PIs.      

\begin{figure}
	\centering
	\includegraphics[scale=0.7]{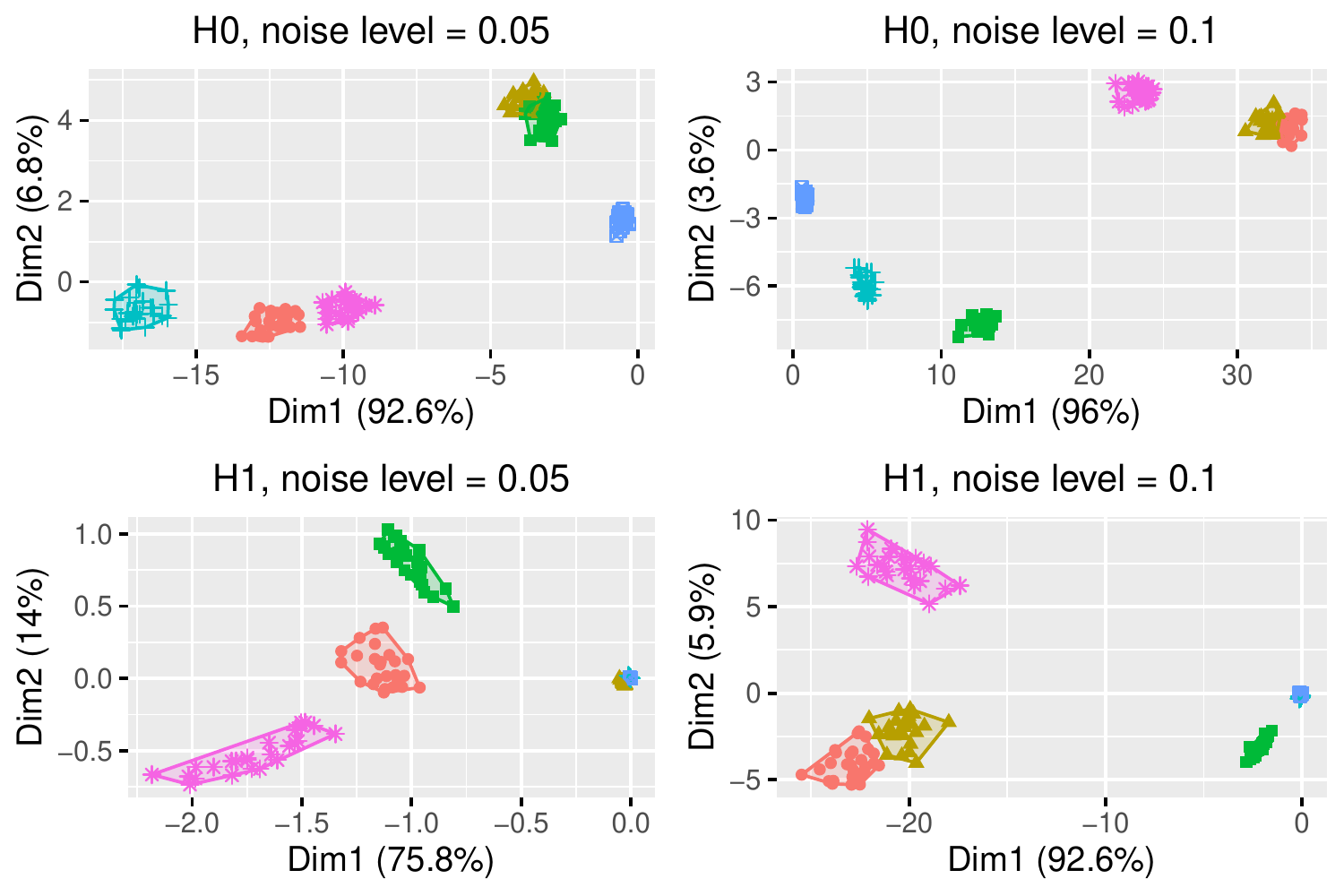}
	\caption{Visualization of VPB-based clusters using PCA.}
	\label{fig:Cluster-visualization}
\end{figure}

Cluster visualizations of VPBs reduced via principal component analysis (PCA) are given in Figure \ref{fig:Cluster-visualization}.


\begin{figure}
	\centering
	\begin{minipage}[b]{.45\textwidth}
		\centering
		\includegraphics[scale=0.4]{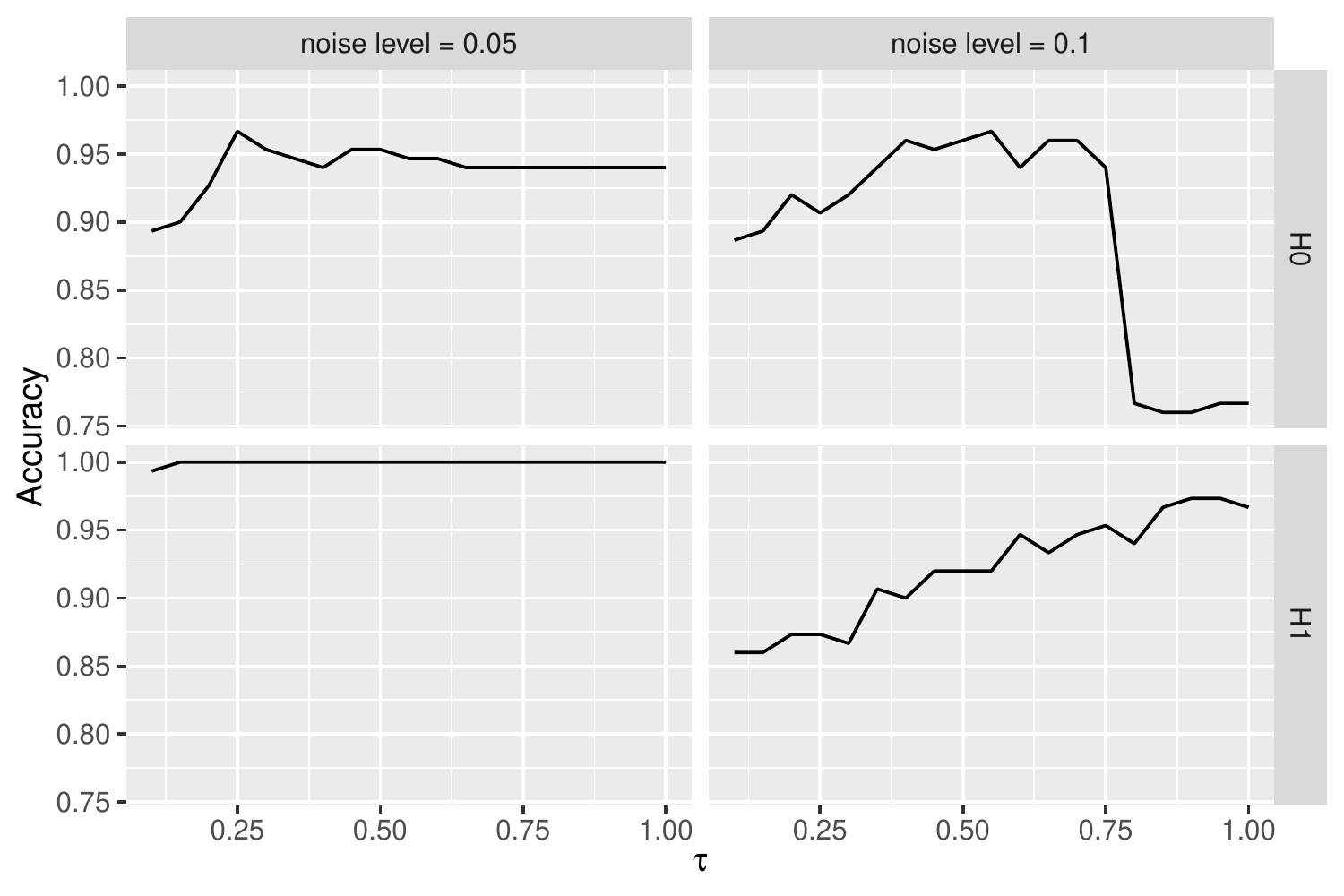}
		\caption{VPB accuracy as a function of $\tau$. Grid size $6 \times 6$, metric $L_2$.}
		\label{fig:sens1}
	\end{minipage}
	\quad
	\begin{minipage}[b]{.45\textwidth}
	 		\centering
		\includegraphics[scale=0.4]{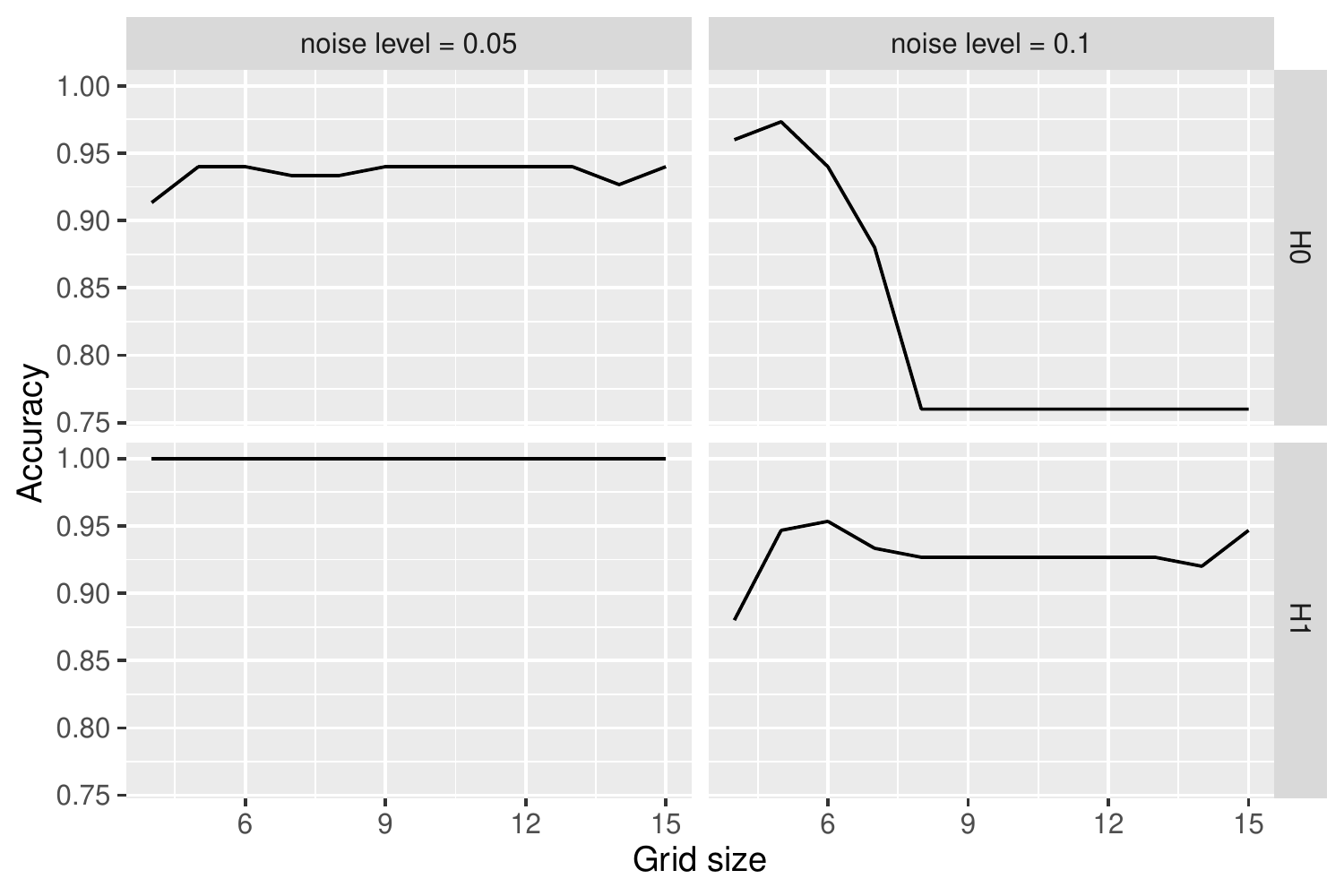}
		\caption{VPB accuracy as a function of grid size. $\tau=0.75$, metric $L_2$.}
		\label{fig:sens2}
	\end{minipage}
\end{figure}

Finally, we explore the sensitivity of accuracies for VPBs with respect to changes in tuning parameters. From Figures \ref{fig:sens1} and \ref{fig:sens2}, we can observe that the accuracies appear to be fairly stable with respect to $\tau$ and grid size when the noise level $\eta=0.05$. 

\subsection{Clustering using K-medoids on a real dataset}

\begin{figure}
	\centering
		\begin{minipage}[b]{.45\textwidth}
		\centering
		\includegraphics[scale=0.4]{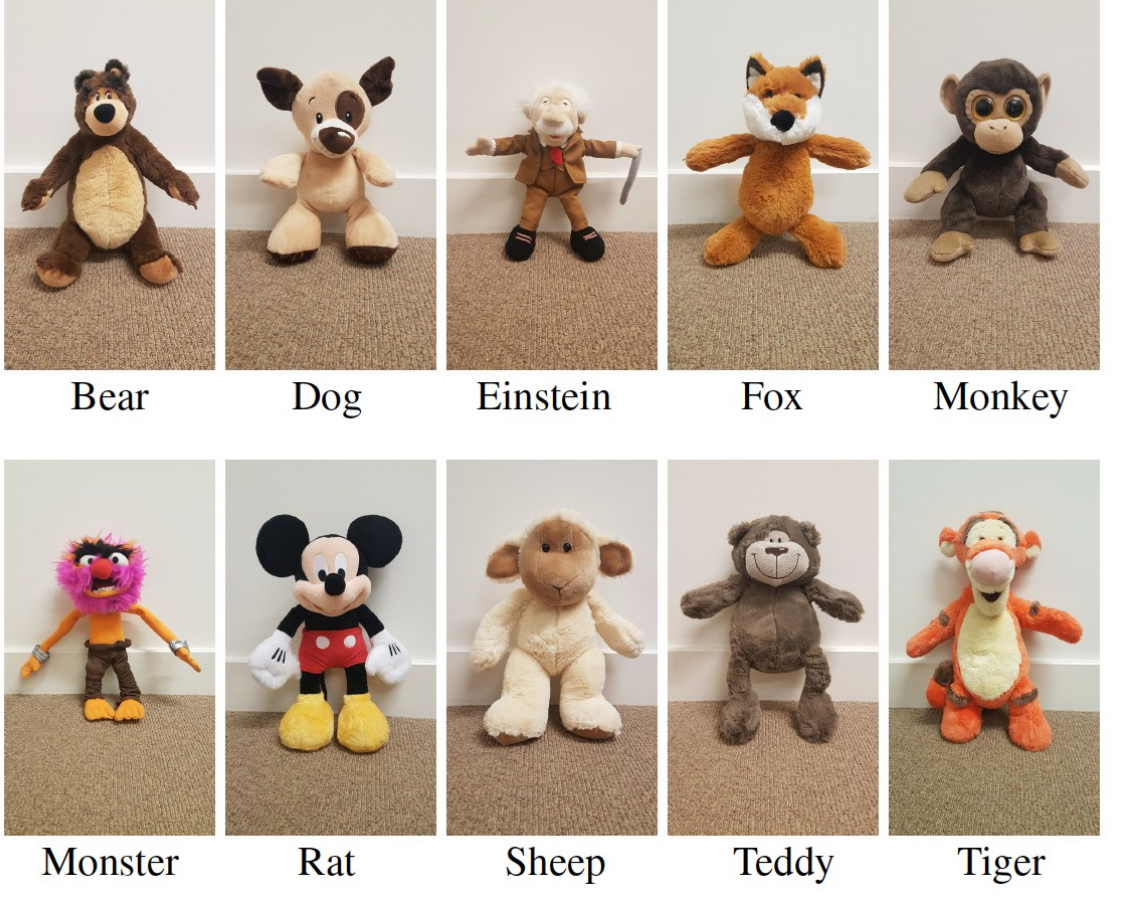}
		\caption{Toys used for SHREC17 dataset \cite{SHREC17}.}
		\label{fig1:SHREC17}
	\end{minipage}
\quad
	\begin{minipage}[b]{.45\textwidth}
	 \centering
	\includegraphics[scale=0.16]{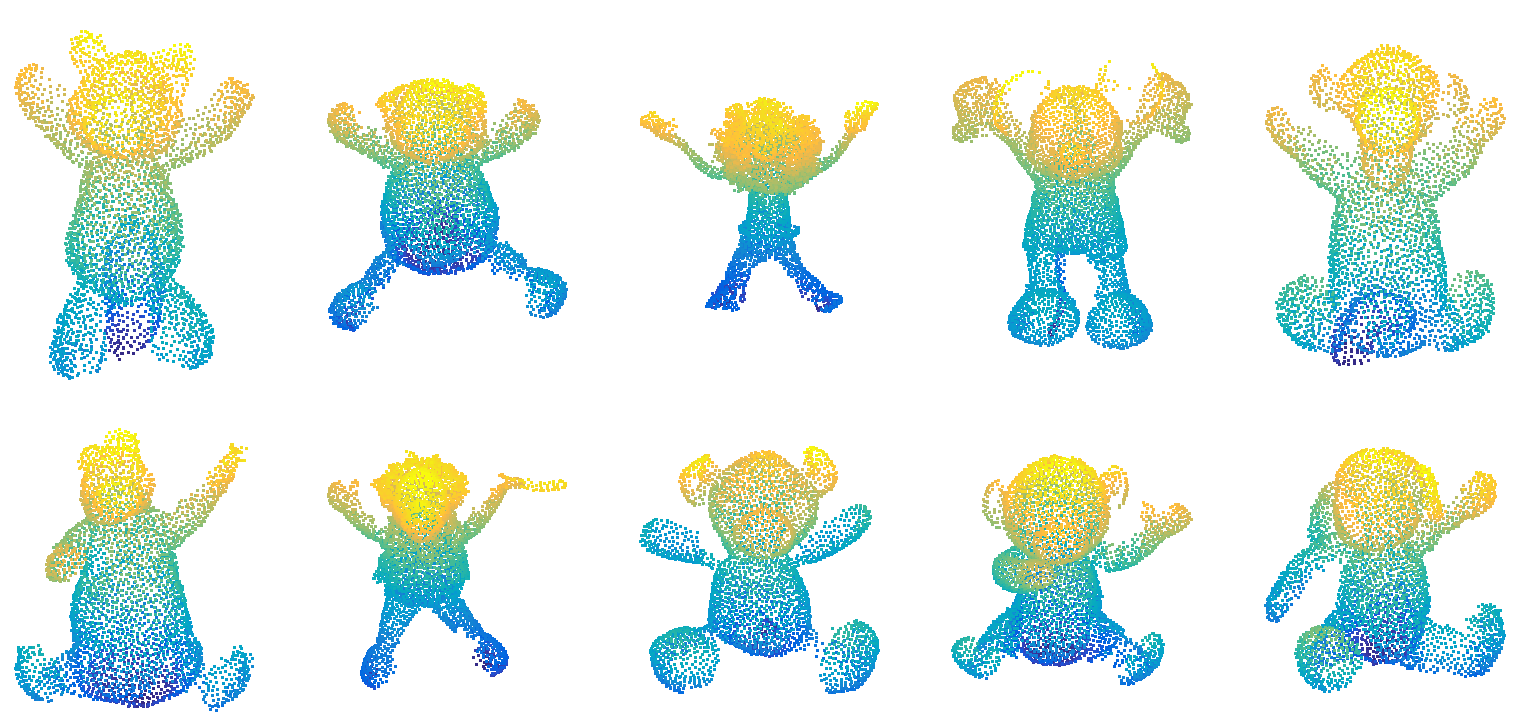}
	\caption{Sample point clouds from SHREC17 dataset. The points are colored based on the $y$ and $z$ coordinates \cite{SHREC17}.}
	\label{fig2:SHREC17}
\end{minipage}
\end{figure}

In our second simulation study, we perform clustering and shape retrieval tasks on a 3D dataset. The dataset consists of 100 point clouds derived from 3D scans of ten non-rigid toys, each having approximately 4000 points \cite{SHREC17}. All toys are scanned in ten different poses by moving various body parts around their joints (see Figures \ref{fig1:SHREC17} and \ref{fig2:SHREC17}). The dataset contains only vertex information and no triangle mesh is provided. The PDs are computed via $\alpha$-shape filtrations built on top of the point clouds.  

\begin{figure}
	\centering
	\includegraphics[scale=0.4]{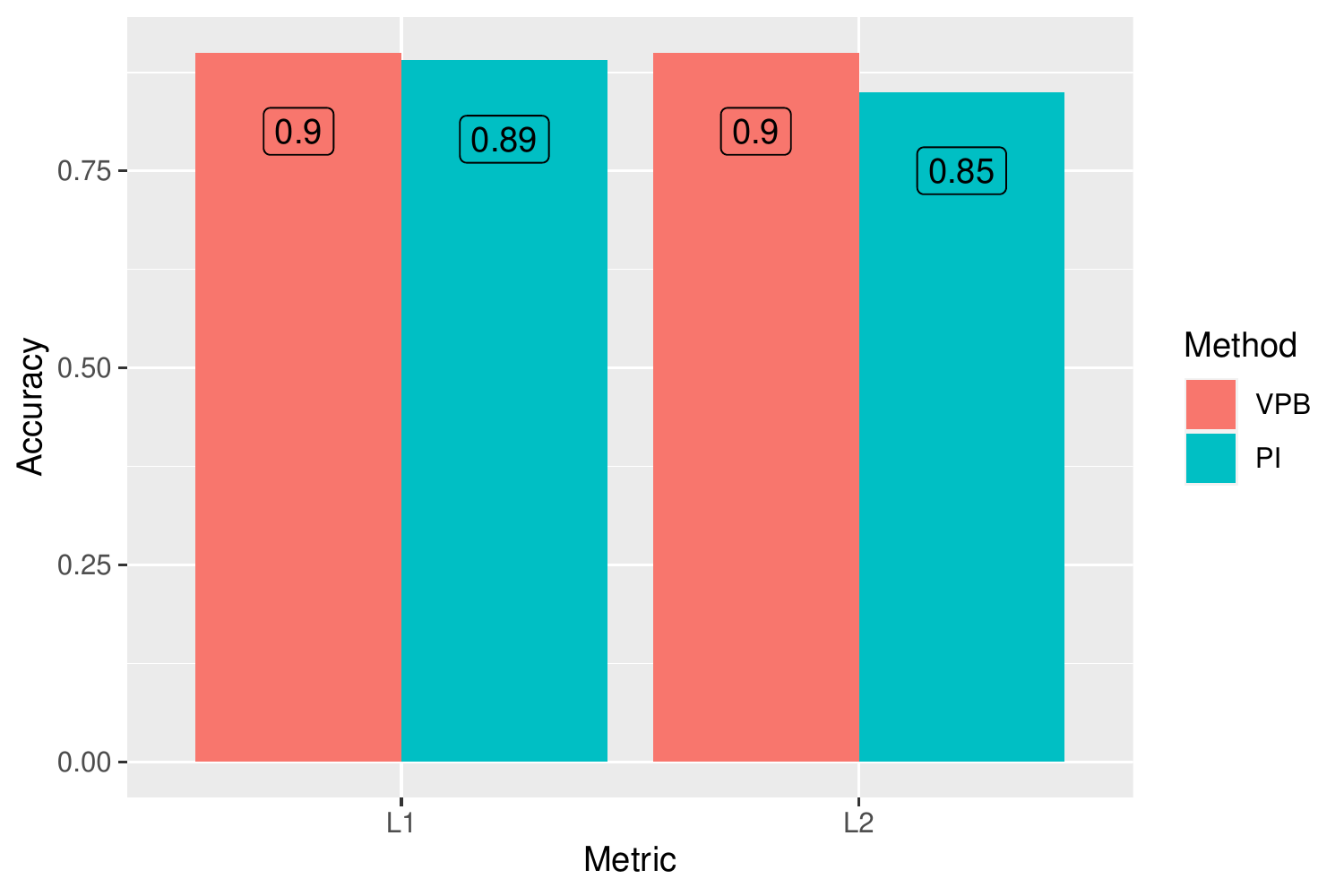}
	\caption{Homological dimension used is $H_2$.}
	\label{fig:SHREC17_acc}
\end{figure}

As in Section \ref{sec1.1}, we perform clustering using the K-medoids algorithm. For PIs, we select a $20 \times 20$ grid and the default value of $\sigma$ which is 1/2 of the maximum persistence across all PDs divided by the grid size. VPBs are computed on a $12\times12$ grid. All the other configurations carry over from the previous section. In particular, similar to the previous experiment, we perform clustering for each homological dimension.
%
%
We find that VPBs and PIs are most discriminative for dimension $H_2$ and therefore we report our results only for this dimension. 
Figure \ref{fig:SHREC17_acc} shows the accuracies achieved by the two methods.


The $100\times100$ dissimilarity matrix used as an input for K-medoids, can also be viewed as a result of a 3D shape retrieval problem. In shape retrieval, as in clustering, no class information is allowed to be used to compute this matrix. We evaluate the retrieval performance using five standard statistics (Nearest Neighbor (NN), First-Tier (FT), Second-Tier (ST), E-Measure (E) and Discounted Cumulative Gain (DCG) \cite{shilane2004princeton}) computed from the dissimilarity matrix. 

The selected dataset for our experiment was initially used for the 2017 Shape Retrieval Contest (SHREC) track: Point-Cloud Shape Retrieval of Non-Rigid Toys. This track invited researchers to contribute to the development of new discriminative and efficient descriptors of 3D non-rigid shapes. A total of eight groups participated in this content. Each group was allowed to submit up to 6 dissimilarity matrices. We compare our results with not only the performance of PIs but also with the best results of all participants. 

\begin{figure}
	\centering
		\begin{tabular}{lccccc}
			\hline
			\hline
			\textbf{Method} & \textbf{NN}& \textbf{FT} & \textbf{ST} & \textbf{E} & \textbf{DCG}\\ 
			\hline
			\hline
			BoW-RoPS-DMF-3 & 1.0000 & 0.9778 & 0.9978 &  0.4390 &  0.9979\\ 
			BPHAPT & 0.9800 & 0.9111 & 0.9544& 0.4273&0.9743\\
			PI & 0.9700	& 0.8989 & 0.9767 &	0.4376 & 0.9740\\
			VPB & 0.9800 & 0.8400 &	0.9567 &	0.4351 & 0.9573\\
			MFLO-FV-IWKS &0.8900 &0.7911& 0.8589 &0.4024 &0.9038\\
			CDSPF & 0.9200 &0.6744 &0.8156 &0.4005 &0.8851\\
			SnapNet &0.8800 &0.6633 &0.8011& 0.3985 & 0.8663 \\
			AlphaVol1 &0.7900 &0.5878 &0.7578& 0.3980 &0.8145\\
			SQFD(WKS) &0.5400 &0.3111 & 0.4467 &0.2507 & 0.6032\\
			m3DSH-2 &0.4400 &0.1867& 0.2856& 0.1932 &0.4997\\
			\hline
		\end{tabular}
		\caption{Retrieval results on SHREC17 \cite{shilane2004princeton}. The methods are ranked according to overall performance.}
		\label{retrieval-results}
\end{figure}

We use $L_1$ metric to compute the dissimilarity matrices for VPBs and PIs in homological dimension $H_2$. For VPBs, $\tau=0.1$ gives a slightly better result than that of $\tau=0.3$ for which the Davies-Bouldin Index is minimum. From Figure \ref{retrieval-results}, we see that PIs slightly outperform VPBs. BPHAPT and our method show comparable results - we have the same NN score and surpass in terms of ST and EE scores.

Furthermore, VPBs exhibit substantial computational advantages over both BoW-RoPS-DMF-3 and BPHAPT. For example, for BoW-RoPS-DMF-3 which utilizes the Bag-of-Words scheme, it takes around 30 minutes for the codebook module to run, whereas our method needs only about 48 seconds to compute all topological summaries (46 seconds for PDs and 2 seconds for VPBs). In terms of data preprocessing, BPHAPT has an extra step of constructing triangular meshes using the Poisson reconstruction method, while our approach does not preprocess the data and only uses the vertex information. Finally, our method has simpler complexity with regard to the number of tuning parameters involved - we only need to choose the parameter $\tau$ and the grid size.

\subsection{Classification of parameter values of a discrete dynamical system}

In this section, we explore the utility of VPBs in the context of classification and change point detection problems. These studies are also motivated by \cite{PI} and involve data generated by the linked twist map defined by the dynamical system:
	\vspace{-0.1cm}
\begin{align*} 
	x_{n+1} &= x_n +r\cdot y_n(1-y_n) \hbox{ mod 1}\\
	y_{n+1} &= y_n +r\cdot x_{n+1}(1-x_{n+1}) \hbox{ mod 1,}
\end{align*}

where $r>0$ \cite{hertzsch2007dna}. As in \cite{PI}, we consider values of $r$ ranging in  $\{2.0,3.5,4.0,4.1,4.3\}$, and take $\sigma=0.005$ and a $20\times20$ grid for the computation of PIs. 

\begin{figure}
 	\centering
 	\begin{minipage}[b]{.45\textwidth}
 		\centering
 		\includegraphics[scale=0.4]{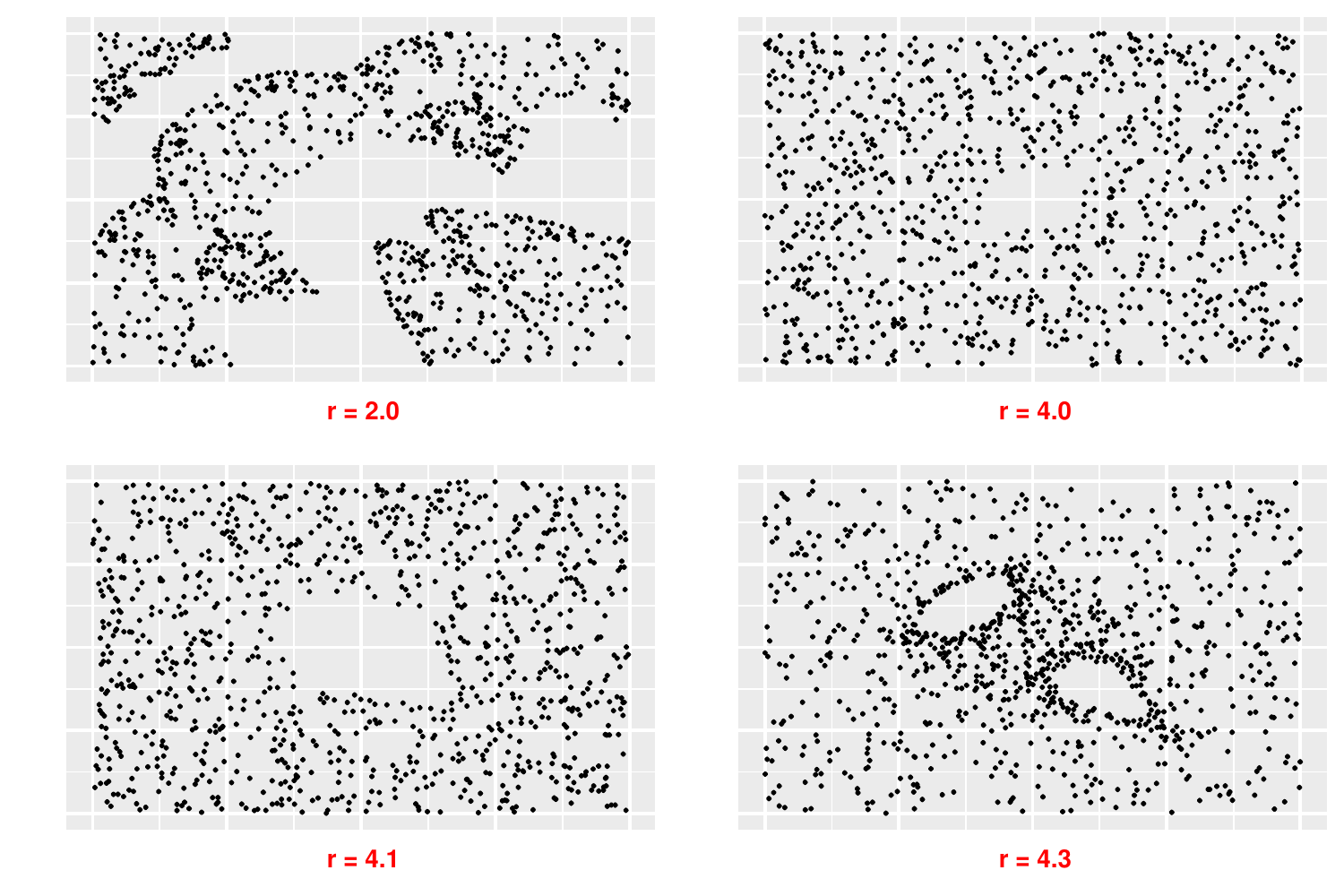}
 		\caption{Scatter plots of sample truncated orbits corresponding to $r=2.0,4.0,4.1,4.3$ respectively.}
 		\label{fig:orbits1-a}
 	\end{minipage}
 	\quad
 	\begin{minipage}[b]{.45\textwidth}
 	 \centering
 	  		\centering
 		\includegraphics[scale=0.4]{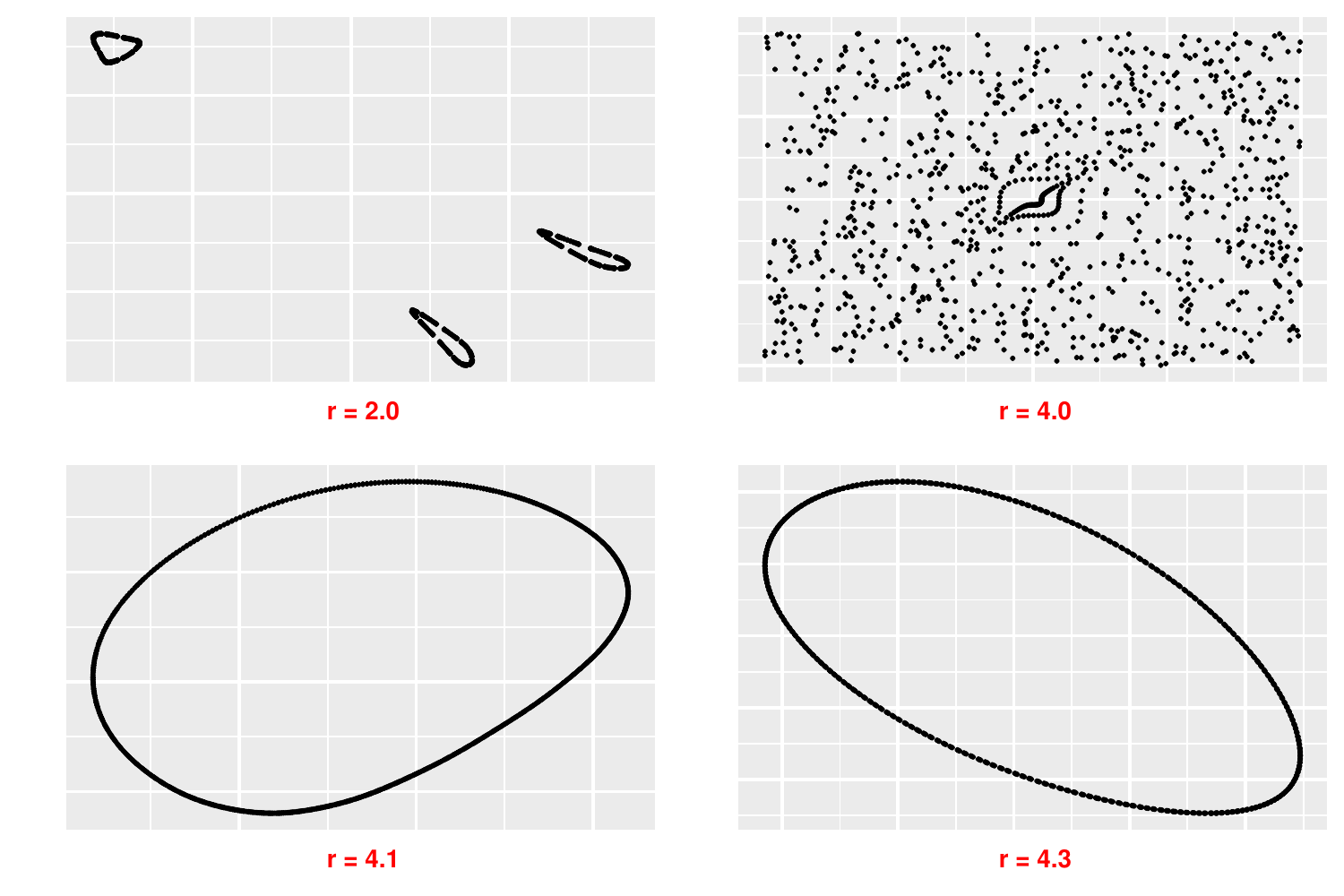}
 		\caption{Scatter plots of atypical orbits corresponding to $r=2.0,4.0,4.1,4.3$ respectively.}
 		\label{fig:orbits2-a}
 	\end{minipage}
 \end{figure}

\begin{figure}
 	\centering
 	\begin{minipage}[b]{.45\textwidth}
 		\centering
 		\includegraphics[scale=0.4]{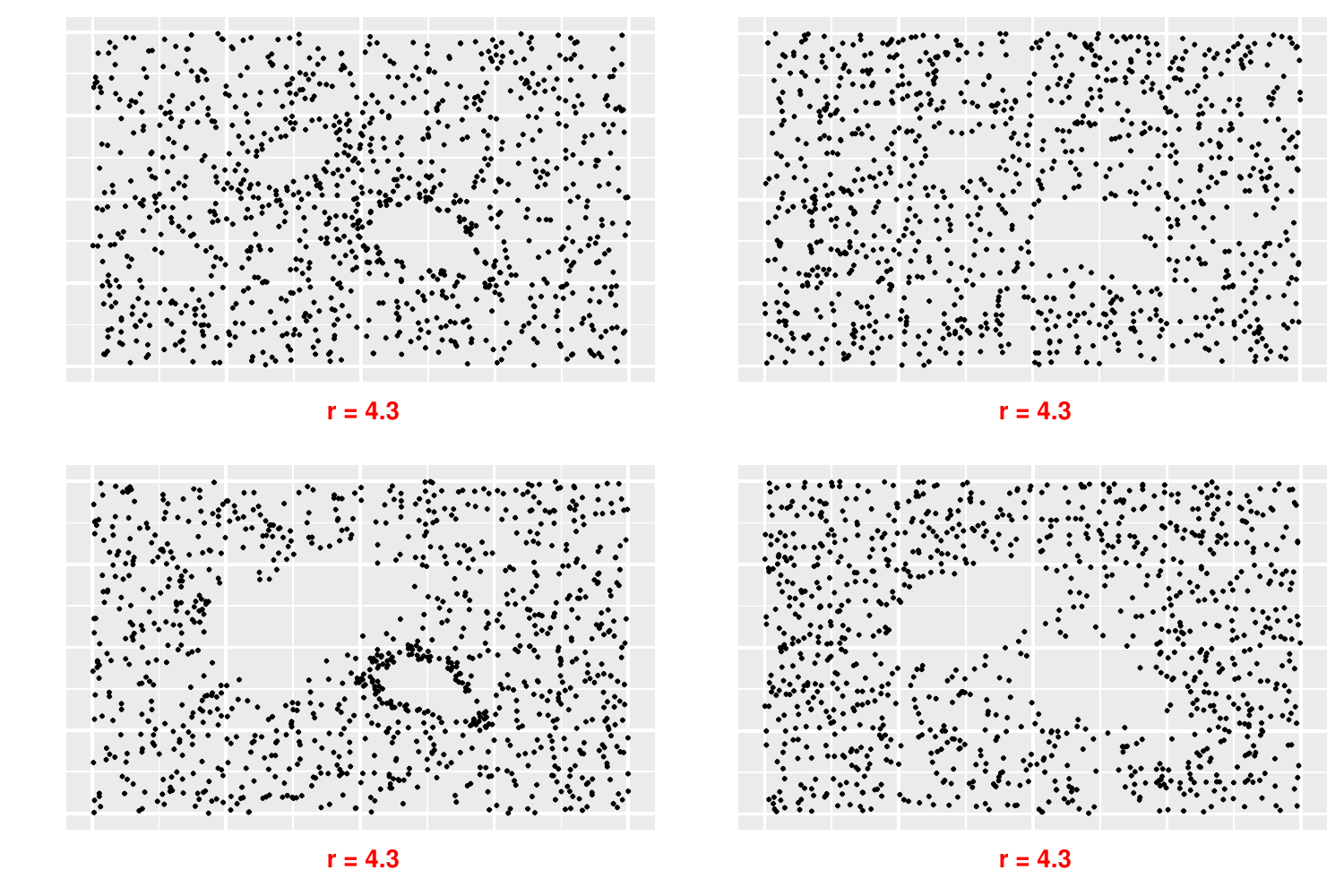}
 		\caption{Scatter plots of sample truncated orbits corresponding to $r=4.3$ and different $(x_0,y)$.}
 		\label{fig:orbits1-b}
 	\end{minipage}
 	\quad
 	\begin{minipage}[b]{.45\textwidth}
 	 		\centering
 		\includegraphics[scale=0.4]{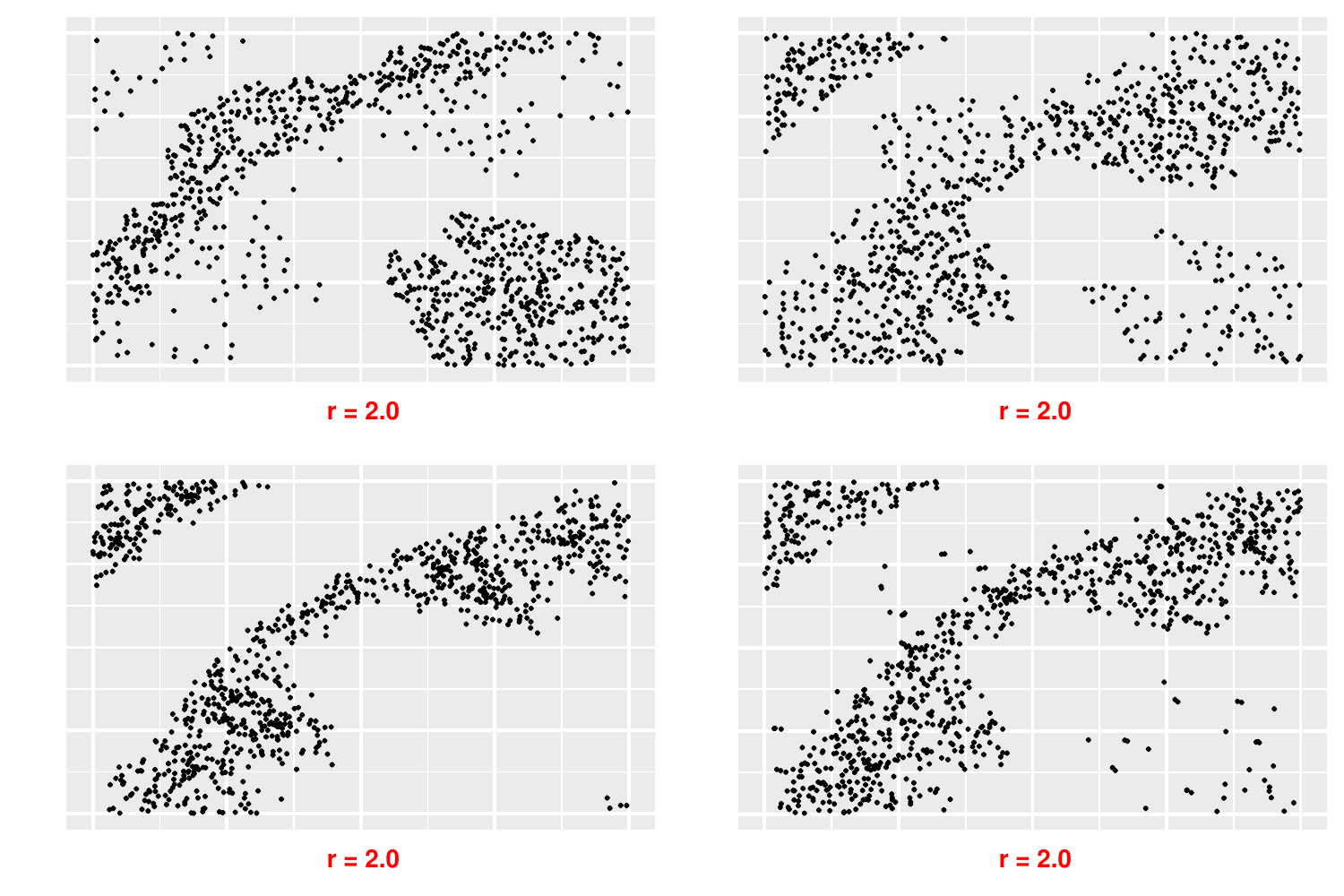}
 		\caption{Scatter plots of sample truncated orbits corresponding to $r=2.0$ and different $(x_0,y)$.}
 		\label{fig:orbits2-b}
 	\end{minipage}
 \end{figure}

For classification, we generate 100 truncated orbits $\{(x_n,y_n):n=0,1,\ldots,1000\}$ for each value of $r$, where the initial points $(x_0,y_0)$ are selected at random from the unit square (see Figures \ref{fig:orbits1-a}, \ref{fig:orbits1-b} and \ref{fig:orbits2-b}; a typical scatter plot for $r=3.5$, not shown in the figures due to space limitations, does not exhibit any pattern and appears to be uniformly distributed over the unit square). The figures show that the scatter plots for $r=4.0$ and $r=4.1$ both have one hole in the center but they slightly differ in size and that scatter plots for the same value of $r$ can show noticeable within-class variation in shape making the classification task more challenging. 

Vietoris-Rips filtrations are used to compute PDs. To curb the computation cost to produce the PDs, we terminate the filtrations at scale value of 0.45. In other words, simplices with at least one pairwise distance greater than 0.45 do not enter any complex in the filtration. At this scale value, the holes in the point clouds (for $r=4.0,4.1,4.3$) are already filled by 2-simplices (triangles). Going beyond that scale value does not essentially produce any additional useful topological information.

Interestingly, when generating the orbits for our simulation studies we notice that for certain values of $r$, some initial points $(x_0,y_0)$ lead to atypical-looking orbits (see Figure \ref{fig:orbits2-a}). Clearly, these orbits would mislead any classification method, so we devise a simple heuristic to omit such orbits from our analysis. 
We compute VPBs on a $7 \times 7$ grid and let $\tau$ vary over $\{0.1,0.3,0.5,0.7,0.9\}$. The optimal value of $\tau$ is selected by 10-fold cross-validation. We use standard machine learning methods such as Random Forest, Support Vector Machines (SVM), GLMNet and Model Averaged Neural Network to classify the orbits in the test set according to the associated parameter value of $r$. 

\begin{figure}
 	\centering
 	\begin{minipage}[b]{.45\textwidth}
 		\centering
 		\includegraphics[scale=0.4]{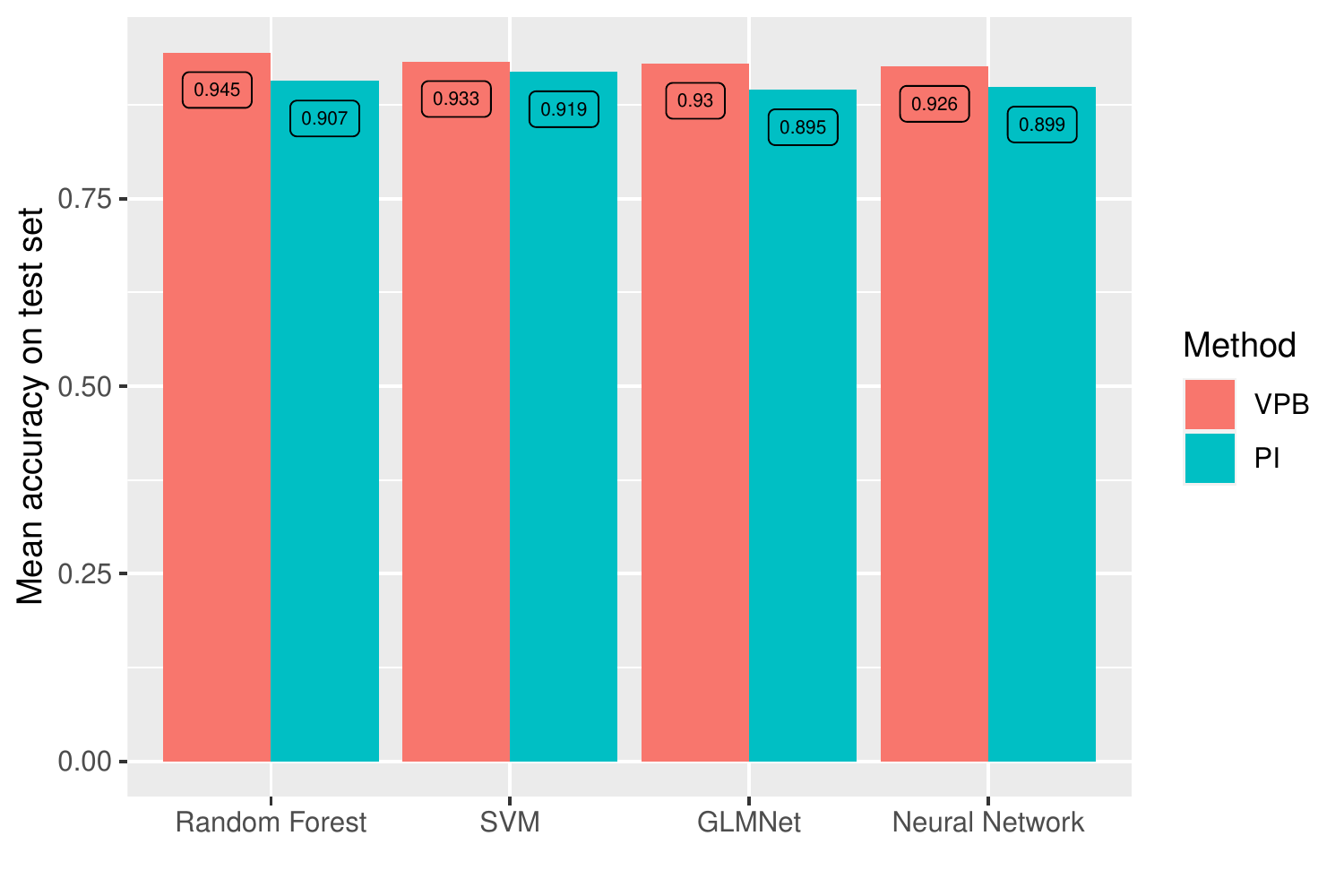}
 		\caption{50/50 split of data. VPBs and PIs are constructed from $H_1$ features.}
 		\label{fig:LTM5050}
 	\end{minipage}
 	\quad
 	\begin{minipage}[b]{.45\textwidth}
 	 	\centering
 		\includegraphics[scale=0.4]{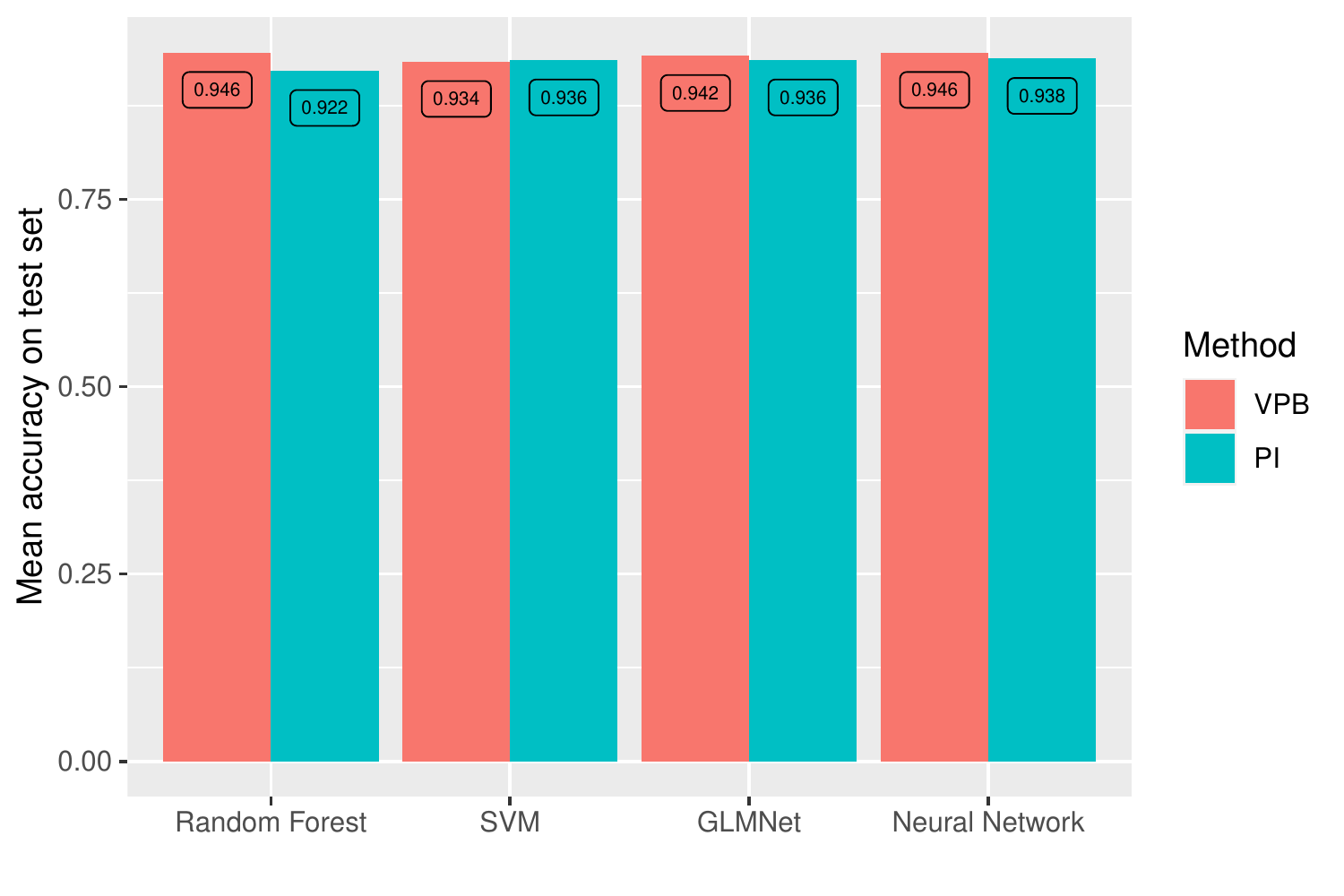}
 		\caption{70/30 split of data. VPBs and PIs are constructed from $H_1$ features.}
 		\label{fig:LTM7030}
 	\end{minipage}
 \end{figure}

We run the classifiers on the $H_0$ and $H_1$ features as well as when these features are concatenated. $H_1$-based test accuracies for 50/50 and 70/30 splits of the data and averaged over 10 independent trials are given in Figures \ref{fig:LTM5050} and \ref{fig:LTM7030}. When the $H_0$ and $H_1$ are concatenated, the accuracies come out to be only about 1\% higher than those for $H_1$, while the $H_0$ features yield the lowest results (around 62\% accuracy). From the figures we can see that, VPBs, when compared to PIs, on average produce slightly higher test accuracies across almost all the machine learning methods employed. 

Next, we present a study on change point detection. Let $r_1=2$, $r_2=3.5$, $r_3=4.0$, $r_4=4.1$, $r_5=4.3$. In this study, we generate 50 truncated orbits for each value of $r$ using the linked twist map and assume they are ordered in time so that the change points occur at times $t=51,101,151,201$: 
$X_{t,r}=\{(x_n(r),y_n(r)):n=0,1,\ldots,1000\}$, where $r=r_{\lceil t/50\rceil}$, $t=1,\ldots,250$ and initial  conditions $(x_0,y_0)$ are selected at random. To estimate the true change points, we use a nonparametric multiple change point detection method called E-Divisive which determines statistically significant change points based on a permutation test \cite{matteson2014nonparametric}. Our choice is motivated by the fact that the distribution of the topological features remains unknown. We input $VPB_{t,r}$ and $PI_{t,r}$ associated with $X_{t,r}$ to {\tt e.divisive()} function from {\tt ecp} package which outputs a list of estimated change points \cite{james2013ecp}. The change point results are given in Figures \ref{CPD1} and \ref{CPD2}, which contain the mean absolute errors computed over 10 independent trials. In case the expected change point is not detected, we set the absolute error equal to 50 - the gap between any two consecutive true change points.

 \begin{figure}
 	\centering
 	\begin{minipage}[b]{.45\textwidth}
	\centering
	\begin{tabular}{lccccc}
		\hline
		& 1st CP & 2nd CP & 3rd CP & 4th CP\\ 
		\hline
		VPB & 0.0 & 43.6 & 7.4 & 33.5 \\ 
		PI & 0.0 & 10.5 & 3.6 & 19.2  \\ 
		\hline
	\end{tabular}
	\caption{Mean absolute errors for $H_0$.}
	\label{CPD1}

 	\end{minipage}
 	\quad
 	\begin{minipage}[b]{.45\textwidth}
	\centering
	\begin{tabular}{lccccc}
		\hline
		& 1st CP & 2nd CP & 3rd CP & 4th CP\\ 
		\hline
		VPB & 0.0 & 0.5 & 1.7 & 1.0 \\ 
		PI & 0.0 & 17.7 & 1.6 & 6.6\\ 
		\hline
	\end{tabular}
	\caption{Mean absolute errors for $H_1$.}
	\label{CPD2}
 	\end{minipage}
 \end{figure} 

Finally, we examine the optimal values of the turning parameter $\tau$ that should have been selected based on the ground truth. We find that if there is little to no noise in the data, then $\tau=0.3$ was optimal for homological dimensions $H_1$ and $H_2$ (when relevant) across all our simulations on clustering and classification. Moreover, we observe that all the classifiers identify $\tau=0.3$ as optimal based on 10-fold cross-validation as well.


%% file: Conclusion.tex
The induced persistence block map presents a way to transform a PD into a vector in the Hilbert space $L^2(wdA)$ where the area measure $wdA$ is induced by a continuous nonnegative weight function $w$. We establish the necessary and sufficient condition for the induced persistence block map to be continuous with respect to the $p$-Wasserstein distance. As a special case, when the persistence block map is induced by the constant function 1, we prove additional results on stability. A VPB is a finite dimensional vector summary of the image of the induced persistence block map for a given PD. Being vectors of $\mathbb{R}^n$, VPBs can easily be used as inputs within many machine learning frameworks. VPBs are faster to compute compared to other vector-based summaries such as persistence images and persistence landscapes. The parameters involved in the computation of VPBs are scale-free which makes their selection or tuning easier in practice. Through several simulation studies covering various learning task such as clustering, classification and change point detection, we have shown that VPBs produce improved performance results over persistence images in most cases.

In our simulation studies, VPBs are derived from persistence blocks induced by the constant function 1. In the future, we plan to consider a wider class of such functions and investigate the corresponding stability properties.

%% file: arxiv_version.bbl
\begin{thebibliography}{10}

\bibitem{edelsbrunner2010computational}
Herbert Edelsbrunner and John Harer.
\newblock {\em Computational topology: an introduction}.
\newblock American Mathematical Soc., 2010.

\bibitem{carlsson2009topology}
Gunnar Carlsson.
\newblock Topology and data.
\newblock {\em Bulletin of the American Mathematical Society}, 46(2):255--308,
  2009.

\bibitem{zomorodian2005computing}
Afra Zomorodian and Gunnar Carlsson.
\newblock Computing persistent homology.
\newblock {\em Discrete \& Computational Geometry}, 33(2):249--274, 2005.

\bibitem{edelsbrunner2008persistent}
Herbert Edelsbrunner, John Harer, et~al.
\newblock Persistent homology-a survey.
\newblock {\em Contemporary mathematics}, 453:257--282, 2008.

\bibitem{rotman2013introduction}
Joseph~J Rotman.
\newblock {\em An introduction to algebraic topology}, volume 119.
\newblock Springer Science \& Business Media, 2013.

\bibitem{kaczynski2006computational}
Tomasz Kaczynski, Konstantin Mischaikow, and Marian Mrozek.
\newblock {\em Computational homology}, volume 157.
\newblock Springer Science \& Business Media, 2006.

\bibitem{mileyko2011probability}
Yuriy Mileyko, Sayan Mukherjee, and John Harer.
\newblock Probability measures on the space of persistence diagrams.
\newblock {\em Inverse Problems}, 27(12):124007, 2011.

\bibitem{cohen2007stability}
David Cohen-Steiner, Herbert Edelsbrunner, and John Harer.
\newblock Stability of persistence diagrams.
\newblock {\em Discrete \& computational geometry}, 37(1):103--120, 2007.

\bibitem{chazal2014persistence}
Fr{\'e}d{\'e}ric Chazal, Vin De~Silva, and Steve Oudot.
\newblock Persistence stability for geometric complexes.
\newblock {\em Geometriae Dedicata}, 173(1):193--214, 2014.

\bibitem{cohen2010lipschitz}
David Cohen-Steiner, Herbert Edelsbrunner, John Harer, and Yuriy Mileyko.
\newblock Lipschitz functions have l p-stable persistence.
\newblock {\em Foundations of computational mathematics}, 10(2):127--139, 2010.

\bibitem{bubenik2018topological}
Peter Bubenik and Tane Vergili.
\newblock Topological spaces of persistence modules and their properties.
\newblock {\em Journal of Applied and Computational Topology}, 2(3):233--269,
  2018.

\bibitem{chen2015statistical}
Yen-Chi Chen, Daren Wang, Alessandro Rinaldo, and Larry Wasserman.
\newblock Statistical analysis of persistence intensity functions.
\newblock {\em arXiv preprint arXiv:1510.02502}, 2015.

\bibitem{kusano2016persistence}
Genki Kusano, Yasuaki Hiraoka, and Kenji Fukumizu.
\newblock Persistence weighted gaussian kernel for topological data analysis.
\newblock In {\em International Conference on Machine Learning}, pages
  2004--2013. PMLR, 2016.

\bibitem{li2014persistence}
Chunyuan Li, Maks Ovsjanikov, and Frederic Chazal.
\newblock Persistence-based structural recognition.
\newblock In {\em Proceedings of the IEEE Conference on Computer Vision and
  Pattern Recognition}, pages 1995--2002, 2014.

\bibitem{reininghaus2015stable}
Jan Reininghaus, Stefan Huber, Ulrich Bauer, and Roland Kwitt.
\newblock A stable multi-scale kernel for topological machine learning.
\newblock In {\em Proceedings of the IEEE conference on computer vision and
  pattern recognition}, pages 4741--4748, 2015.

\bibitem{bubenik2015statistical}
Peter Bubenik.
\newblock Statistical topological data analysis using persistence landscapes.
\newblock {\em The Journal of Machine Learning Research}, 16(1):77--102, 2015.

\bibitem{rieck2017topological}
Bastian Rieck, Filip Sadlo, and Heike Leitte.
\newblock Topological machine learning with persistence indicator functions.
\newblock In {\em Topological Methods in Data Analysis and Visualization},
  pages 87--101. Springer, 2017.

\bibitem{PI}
Henry Adams, Tegan Emerson, Michael Kirby, Rachel Neville, Chris Peterson,
  Patrick Shipman, Sofya Chepushtanova, Eric Hanson, Francis Motta, and Lori
  Ziegelmeier.
\newblock Persistence images: A stable vector representation of persistent
  homology.
\newblock {\em The Journal of Machine Learning Research}, 18(1):218--252, 2017.

\bibitem{berry2020functional}
Eric Berry, Yen-Chi Chen, Jessi Cisewski-Kehe, and Brittany~Terese Fasy.
\newblock Functional summaries of persistence diagrams.
\newblock {\em Journal of Applied and Computational Topology}, 4(2):211--262,
  2020.

\bibitem{atienza2020stability}
Nieves Atienza, Roc{\'\i}o Gonz{\'a}lez-D{\'\i}az, and Manuel
  Soriano-Trigueros.
\newblock On the stability of persistent entropy and new summary functions for
  topological data analysis.
\newblock {\em Pattern Recognition}, 107:107509, 2020.

\bibitem{richardson2014efficient}
Eitan Richardson and Michael Werman.
\newblock Efficient classification using the euler characteristic.
\newblock {\em Pattern Recognition Letters}, 49:99--106, 2014.

\bibitem{chung2019persistence}
Yu-Min Chung and Austin Lawson.
\newblock Persistence curves: A canonical framework for summarizing persistence
  diagrams.
\newblock {\em arXiv preprint arXiv:1904.07768}, 2019.

\bibitem{RUDIN}
Walter Rudin.
\newblock {\em Functional Analysis, Second Edition}.
\newblock International Series in Pure and Applied Mathematics. McGraw-Hill,
  1991.

\bibitem{ghrist2008barcodes}
Robert Ghrist.
\newblock Barcodes: the persistent topology of data.
\newblock {\em Bulletin of the American Mathematical Society}, 45(1):61--75,
  2008.

\bibitem{edelsbrunner1994three}
Herbert Edelsbrunner and Ernst~P M{\"u}cke.
\newblock Three-dimensional alpha shapes.
\newblock {\em ACM Transactions on Graphics (TOG)}, 13(1):43--72, 1994.

\bibitem{fasy2021package}
Brittany~T Fasy, Jisu Kim, Fabrizio Lecci, Clement Maria, David~L Millman,
  Vincent Rouvreau, and Maintainer~Jisu Kim.
\newblock Package ‘tda’, 2021.

\bibitem{wadhwa2018tdastats}
Raoul~R Wadhwa, Drew~FK Williamson, Andrew Dhawan, and Jacob~G Scott.
\newblock Tdastats: R pipeline for computing persistent homology in topological
  data analysis.
\newblock {\em Journal of open source software}, 3(28):860, 2018.

\bibitem{kernelTDApackage}
Tullia Padellini, Francesco Palini, Pierpaolo Brutti, Chih-Chung Chang,
  Chih-Chen Lin, Michael Kerber, Dmitriy Morozov, Arnur Nigmetov, and
  Maintainer~Tullia Padellini.
\newblock Statistical learning with kernel for persistence diagrams, 2020.

\bibitem{halkidi2002clustering}
Maria Halkidi, Yannis Batistakis, and Michalis Vazirgiannis.
\newblock Clustering validity checking methods.
\newblock {\em ACM Sigmod Record}, 31(3):19--27, 2002.

\bibitem{rdusseeun1987clustering}
LKPJ Rdusseeun and P~Kaufman.
\newblock Clustering by means of medoids.
\newblock In {\em Proceedings of the Statistical Data Analysis Based on the L1
  Norm Conference, Neuchatel, Switzerland}, pages 405--416, 1987.

\bibitem{park2009simple}
Hae-Sang Park and Chi-Hyuck Jun.
\newblock A simple and fast algorithm for k-medoids clustering.
\newblock {\em Expert systems with applications}, 36(2):3336--3341, 2009.

\bibitem{lloyd1982least}
Stuart Lloyd.
\newblock Least squares quantization in pcm.
\newblock {\em IEEE transactions on information theory}, 28(2):129--137, 1982.

\bibitem{clusterRpackage}
Martin Maechler, Peter Rousseeuw, Anja Struyf, Mia Hubert, and Kurt Hornik.
\newblock {\em cluster: Cluster Analysis Basics and Extensions}, 2021.
\newblock R package version 2.1.2 --- For new features, see the 'Changelog'
  file (in the package source).

\bibitem{SHREC17}
Frederico~A Limberger, Richard~C Wilson, M~Aono, N~Audebert, A~Boulch,
  B~Bustos, A~Giachetti, A~Godil, B~Le~Saux, B~Li, et~al.
\newblock Shrec'17 track: Point-cloud shape retrieval of non-rigid toys.
\newblock In {\em 10th Eurographics workshop on 3D Object retrieval}, pages
  1--11, 2017.

\bibitem{shilane2004princeton}
Philip Shilane, Patrick Min, Michael Kazhdan, and Thomas Funkhouser.
\newblock The princeton shape benchmark.
\newblock In {\em Proceedings Shape Modeling Applications, 2004.}, pages
  167--178. IEEE, 2004.

\bibitem{hertzsch2007dna}
Jan-Martin Hertzsch, Rob Sturman, and Stephen Wiggins.
\newblock Dna microarrays: design principles for maximizing ergodic, chaotic
  mixing.
\newblock {\em Small}, 3(2):202--218, 2007.

\bibitem{matteson2014nonparametric}
David~S Matteson and Nicholas~A James.
\newblock A nonparametric approach for multiple change point analysis of
  multivariate data.
\newblock {\em Journal of the American Statistical Association},
  109(505):334--345, 2014.

\bibitem{james2013ecp}
Nicholas~A James and David~S Matteson.
\newblock ecp: An r package for nonparametric multiple change point analysis of
  multivariate data.
\newblock {\em arXiv preprint arXiv:1309.3295}, 2013.

\end{thebibliography}
